\documentclass[lettersize,journal]{IEEEtran}
\usepackage{amsmath,amsfonts}
\usepackage[linesnumbered, ruled]{algorithm2e}
\SetKwRepeat{Do}{do}{while}%
\usepackage{array}
\usepackage[caption=false,font=normalsize,labelfont=sf,textfont=sf]{subfig}
\usepackage{textcomp}
\usepackage{booktabs, multicol, multirow}
\usepackage{stfloats}
\usepackage{url}
\usepackage{verbatim}
\usepackage{titlesec}
\usepackage{graphicx}
\usepackage{cite}
\usepackage{bm}

\usepackage{amssymb}
\usepackage{graphicx}
\usepackage{float}
\usepackage{amsmath}
\usepackage[dvipsnames]{xcolor}
\usepackage{colortbl}
\usepackage{bbm}
\usepackage[linesnumbered,ruled]{algorithm2e}
\usepackage{enumitem}

\usepackage{hyperref}
\usepackage{cleveref}
\usepackage{mathtools}
\usepackage{float}
\usepackage[bb=boondox]{mathalfa}
\newcommand{\ones}[1]{\mathbbm{1}_{#1}}
\newcommand{\zeros}[1]{\mathbb{0}_{#1}}

\makeatletter
\DeclareRobustCommand\onedot{\futurelet\@let@token\@onedot}
\def\@onedot{\ifx\@let@token.\else.\null\fi\xspace}
\def\eg{\emph{e.g}\onedot} 
\def\ie{\emph{i.e}\onedot}

\makeatother

\begin{document}
\title{A Unified Optimal Transport Framework for Cross-Modal Retrieval with Noisy Labels}

\author{Haochen Han, Minnan Luo, Huan Liu, Fang Nan
\thanks{
This work is supported by the National Key Research and Development Program of China (No. 2022YFB3102600), National Nature Science Foundation of China (No. 62192781, No. 62272374, No. 62202367, No. 62250009, No. 62137002), Project of China Knowledge Center for Engineering Science and Technology,  Project of Chinese academy of engineering ``The Online and Offline Mixed Educational Service System for `The Belt and Road' Training in MOOC China'', and the K. C. Wong Education Foundation.

Haochen Han, Minnan Luo, and Huan Liu are with the School of Computer Science and Technology, Xi’an Jiaotong University, Xi’an 710049, China (E-mail: hhc1997@stu.xjtu.edu.cn; minnluo@xjtu.edu.cn; huanliu@xjtu.edu.cn). 

Fang Nan is with the School of Cyber Science and Engineering, Xi’an Jiaotong University, Xi’an, 710049, China. (E-mail: nanfangalan@gmail.com).

(\textit{Corresponding author: Minnan Luo})
}

}

\maketitle
\IEEEpubidadjcol
\IEEEpubid{\begin{minipage}{\textwidth}\ \\[12pt] \centering
  0000--0000/00\$00.00~\copyright~2021 IEEE\\
\end{minipage}}

\begin{abstract}
Cross-modal retrieval (CMR) aims to establish interaction between different modalities, among which supervised CMR is emerging due to its flexibility in learning semantic category discrimination. Despite the remarkable performance of previous supervised CMR methods, much of their success can be attributed to the well-annotated data. However, even for unimodal data, precise annotation is expensive and time-consuming, and it becomes more challenging with the multimodal scenario. In practice, massive multimodal data are collected from the Internet with coarse annotation, which inevitably introduces noisy labels. Training with such misleading labels would bring two key challenges---enforcing the multimodal samples to \emph{align incorrect semantics} and \emph{widen the heterogeneous gap}, resulting in poor retrieval performance. To tackle these challenges, this work proposes UOT-RCL, a Unified framework based on Optimal Transport (OT) for Robust Cross-modal Retrieval. First, we propose a semantic alignment based on partial OT to progressively correct the noisy labels, where a novel cross-modal consistent cost function is designed to blend different modalities and provide precise transport cost. Second, to narrow the discrepancy in multi-modal data, an OT-based relation alignment is proposed to infer the semantic-level cross-modal matching. Both of these two components leverage the inherent correlation among multi-modal data to facilitate effective cost function. The experiments on three widely-used cross-modal retrieval datasets demonstrate that our UOT-RCL surpasses the state-of-the-art approaches and significantly improves the robustness against noisy labels.

\end{abstract}


\section{Introduction}
\IEEEPARstart{A}s a key step towards artificial general intelligence, multimodal learning aims to understand and integrate information from multiple sensory modalities to mimic human cognitive activities. Cross-modal retrieval is one of the most fundamental techniques in multimodal learning due to its flexibility in bridging different modalities, which has powered various real-world applications, including audio-visual retrieval \cite{zheng2021adversarial, zhang2023variational}, visual question answering \cite{zhang2023lois,sheng2021human,qian2023locate}, and recipe recommendation \cite{zhu2021learning, wang2021cross}.

Based on retrieval objectives, existing CMR methods can be broadly categorized into two groups: the unsupervised CMR for aligning instance-level cross-modal items and the supervised CMR for aligning category-level ones. Due to the ability to associate multi-modal data with semantic categories, supervised CMR becomes a fundamental technique in the label-aware multimedia community, \eg, medical image query report \cite{zhang2022deep} and product recommendation \cite{ma2022ei}, which is also the research topic in this paper.

\begin{figure}[t]
\centering  
\includegraphics[width=0.9\columnwidth]{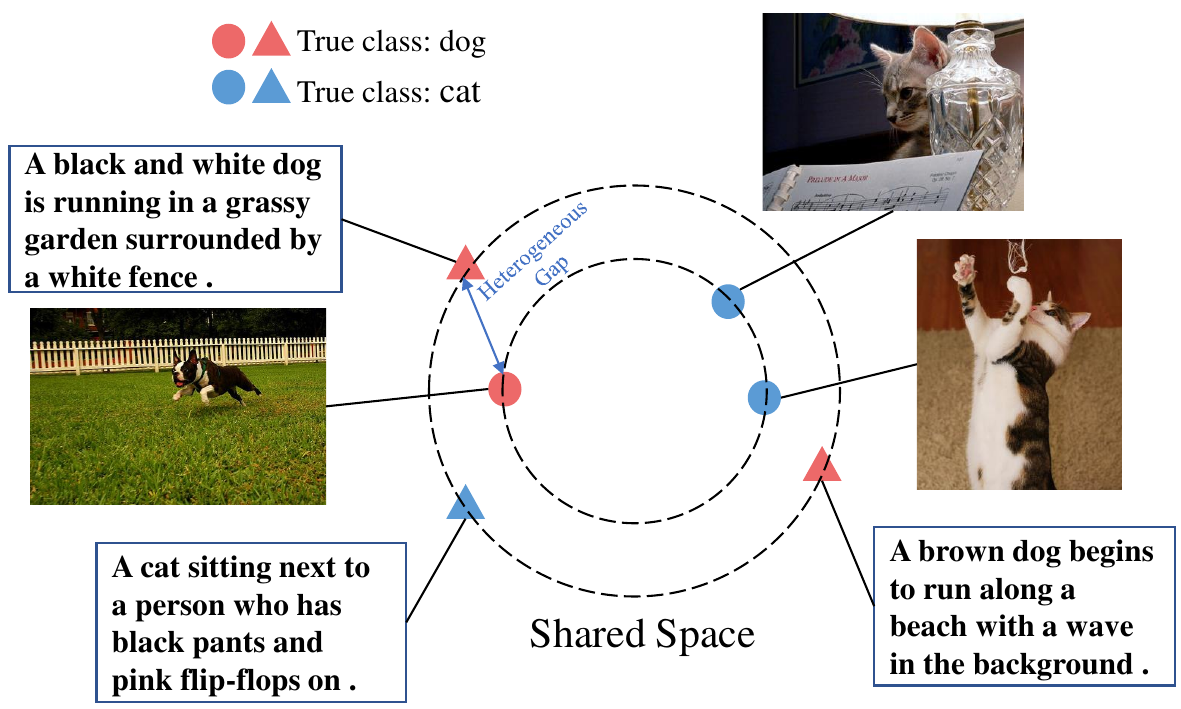}
\caption{Training with noisy labels will result in poor cross-modal retrieval performance. On the one hand, noisy labels can wrongly enforce irrelevant samples to be similar in the shared space. On the other hand, noisy labels can confuse the discriminative connections among different modalities and thus widen the heterogeneous gap.}
\label{fig:intro}
\vspace{-0.3cm} 
\end{figure}

Like many other data-driven tasks, supervised CMR requires massive and high-quality labeled data, which is notoriously labor-intensive.  In practice, even for unimodal scenarios, data labeling can suffer from inherent and pervasive label noise, not to mention the extra obstacles from multiple modalities. Taking the image-text pair as an example, while the visual content is intuitive, the semantic label of the textual description is harder to determine due to its subjectivity. Moreover, different modalities lay in separate regions of the shared space and thus exist vast distinctions, which is called heterogeneity gap \cite{liang2022mind}, making learning discriminative representations more difficult. As shown in Fig.\ref{fig:intro}, training with noisy labels would bring two key challenges that enforce the semantically irrelevant data to be similar and widen the heterogeneous gap, which significantly harms the performance of the CMR model. Therefore, endowing supervised CMR methods with robustness against noisy labels is crucial to suit real-world retrieval scenarios.

To mitigate the impact of noisy labels, most of the existing studies attempt to design the robust loss functions \cite{ma2020normalized,sun2022pnp}, reduce the weight of noisy samples \cite{shu2019meta,ma2021learning}, select clean samples \cite{cheng2021learning,karim2022unicon}, or generate corrected labels \cite{zheng2021meta,li2022devil}. Despite the success in unimodal scenarios, they cannot directly integrate multiple modalities to cope with the noisy labels in the CMR task. To date, efforts that have been made to achieve robust CMR are still limited. Specifically, MRL \cite{hu2021learning} combines robust clustering loss and multimodal contrastive loss to combat noisy samples and correlate distinct modalities simultaneously. Inspired by the memorization effect of DNNs \cite{arpit2017closer}, ELRCMR \cite{xu2022early} introduces an early-learning regularization to further enhance the robustness of clustering and contrasting loss. However, on the one hand, such clustering loss can only down-weight the relative contributions for the incorrect samples, which do not take full advantage of training data. On the other hand, the multimodal contrastive loss only focuses on learning instance-level discriminative representations and neglects the potential semantically relevant data. Thus, the achieved performance by them is argued to be sub-optimal.

To bridge the aforementioned gap, we propose UOT-RCL, a \textbf{U}nified framework based on \textbf{O}ptimal \textbf{T}ransport for \textbf{R}obust \textbf{C}ross-modal \textbf{R}etrieval. Our UOT-RCL performs both the noisy label problem and multi-modal discrepancy from an OT perspective, leveraging the inherent correlation among multi-modal data to construct effective transport cost. First, to alleviate the influence of noisy labels, we view the label correction procedure as a partial OT problem that progressively transports noisy labeled samples into correct semantics at a minimum cost, where a novel cross-modal consistent cost function is proposed to blend different modalities and provide precise transport cost. To ensure the reliability of the corrected label, we gradually increase the transport mass over the course of training. Second, to narrow the heterogeneous gap of multi-modal data, we model an OT problem to infer the semantic-level matching among distinct modalities, where a relation-based cost function is proposed to preserve the connection of training samples and the selected confident samples. The above two OT-based components can be solved using the efficient Sinkhorn-Knopp algorithm \cite{cuturi2013sinkhorn} and summarized in one unified framework.

Our main contributions are summarized as follows:
\begin{itemize}
    \item We propose a novel OT-based framework for supervised CMR with noisy labels, which is a practical but rarely-explored problem in multi-modal learning.
    
    \item We present a semantic alignment based on partial OT to progressively correct the noisy labels, where a novel cross-modal consistent cost function is designed to blend different modalities and provide precise transport cost.
    
    \item An OT-based relation alignment is proposed to infer the semantic-level cross-modal matching, which is used to narrow the heterogeneous gap among modalities.
    
    \item We conduct extensive experiments on three widely used multimodal datasets, which clearly demonstrate the robust performance of our method against noisy labels.
\end{itemize}

\section{Related Works}
In this section, we briefly review three related topics to our study, including cross-modal retrieval, learning with noisy labels, and optimal transport.

\subsection{Cross-Modal Retrieval}
Cross-modal retrieval aims to retrieve relevant samples from different modalities for a given query. Current dominant methods for CMR is to project distinct modalities into a shared feature space to reduce the heterogeneous gap, wherein the key is to measure the cross-modal similarity. 
Based on the usage of annotation and retrieval scenes, these methods could be roughly divided into two groups: 1) Unsupervised methods that directly maximize the co-occurred cross-modal agreement between different modalities. One typical research line is global alignment, aiming at learning the correspondence between entire cross-modal data. Existing studies usually adopt a two-stream network to learn comparable global features \cite{vse++}. Another research line is local alignment which focuses on aligning the local cross-modal regions for more fine-grained CMR. For example, \cite{lee2018stacked, li2019visual} design the cross-attention mechanism to learn alignment between the local regions and words. 2) Supervised methods utilize semantic supervision and aim to constrain the cross-modal instances to be similar for the same category. For example, some methods \cite{hu2019multi,hu2019multimodal} design the discriminative criteria to maximize the intra-class similarity while minimizing the inter-class similarity. Alternatively, some methods\cite{hu2019scalable,zhen2019deep} directly utilize a common classifier to enforce multi-modal networks to learn a shared discriminative space. In addition, adversarial learning-based methods \cite{wang2017adversarial} typically construct an adversarial process to seek the effective common subspace. 

Our proposed approach falls in the category of supervised methods. Despite the success of these prior arts, they heavily rely on well-annotated data and cannot tackle the noise issue. This study aims to find a solution for robust supervised CMR against noisy labels, which is practical but less touched before.

\subsection{Learning with Noisy Labels}
To mitigate the impact of the errors in annotations, numerous approaches have been carried out in past years, which typically consist of sample reweighting, label correction, and noisy sample selection. Sample reweighting methods \cite{ren2018learning,shu2019meta,ma2021learning} aim to assign lower weights to potentially noisy labels to reduce their contribution during training. For example, MW-Net \cite{shu2019meta} learns an explicit loss-weight function in a meta-learning manner. Label correction methods \cite{li2021trustable, chen2023refining} seek to transform noisy labels to correct ones directly, which have shown more favorable results than other methods. DivideMix \cite{li2019dividemix} employs two co-teaching networks to correct targets from the average predictions of different data augmentations. MLC \cite{zheng2021meta} trains a label correction network based on meta-learning to generate corrected labels for training data. Noisy sample selection methods aim to choose and remove noisy labeled data within the training datasets. Existing works are mostly based on the memorization effect of deep networks that samples with higher loss are prone to hold noisy labels and thus exploit the small-loss criterion to divide data. Some recent advances \cite{nishi2021augmentation, yang2022learning} model per-sample loss distribution with mixture probabilistic models to separate noisy samples. Alternatively, some methods \cite{li2022neighborhood, li2022selective} resort to neighborhood information in feature space to identify noisy labels. In addition, some state-of-the-art methods combine several techniques to further improve the robustness of models, \eg, ELR \cite{liu2020early}. However, most prior arts are designed for the unimodal scenario, and it is difficult or even impossible to apply them to multimodal cases. To combat the noisy labels in CMR, MRL \cite{hu2021learning} combines two multimodal robust losses to alleviate the influence of noisy samples. ELRCMR \cite{xu2022early} introduces a regularization based on early learning to prevent the memorization of noisy labels. Differently, our work tackles this challenge from an optimal transport perspective.

\subsection{Optimal Transport}
Optimal transport is proposed to measure the distance between two probability distributions. One of the factors limiting the widespread application of OT is the high computational cost associated with using linear programming solvers. To improve its availability, entropy-regularized OT \cite{cuturi2013sinkhorn} is proposed to provide a computationally cheaper solver. Benefiting from this, OT has drawn much attention in various fields of machine learning, which include but are not limited to unsupervised learning \cite{caron2020unsupervised,asano2020labelling}, semi-supervised learning \cite{tai2021sinkhorn}, object detection \cite{ge2021ota,afouras2022self}, domain adaptation \cite{redko2019optimal,fatras2022optimal}, and long-tailed recognition \cite{peng2022optimal,wang2022solar}.  Recently, OT-Filter \cite{feng2023ot} studies the problem of noisy sample selection from the perspective of OT to combat label noise. In contrast, this paper proposes a unified OT-based framework for robust CMR that simultaneously corrects noisy labels and narrows the heterogeneous gap.

\section{Background}
\subsection{Problem Setup}
Without losing generality, we focus on bimodal data, \ie, visual-text pair, to present the noisy label problem in CMR. We assume that there is a collection of $N$ visual-text pairs $\mathbbm{D}=\{(\bm{x}_i^v, \bm{x}_i^t, \bm{y}_i)\}_{i=1}^N$, where $(\bm{x}_i^v, \bm{x}_i^t)$ is a visual-text pair and $\bm{y}_i \in [0,1]^K $ is the one-hot semantic label over $K$ classes. The goal of supervised CMR is to project the bimodal data into a shared feature space wherein data pairs belonging to the same category have higher feature similarities. Generally, existing methods attempt to learn two modality-specific networks: $\bm{z}_i^v = f_v(\bm{x}_i^v;\Theta_v) \in \mathbb{R}^L$ for visual modality and $\bm{z}_i^t = f_t(\bm{x}_i^t;\Theta_t) \in \mathbb{R}^L$ for text modality, where $L$ is the dimension of shared space, and $\Theta_v$ and $\Theta_t$ are the trainable parameters of the two networks. To align cross-modal samples, existing works typically use $K$ learnable common prototype $\bm{\mu}_k \in \mathbb{R}^L$ corresponding to each class $k \in \{1,2,\ldots,K \}$ to calculate the probability of $i$-th sample belonging to the $k$-th class:
\begin{equation}\label{eq:probability}
	 p(k|\bm{x}_i^v) = 
	  \frac{\exp \left(\bm{\mu}_k^\top \bm{z}_i^v / \tau_1\right)}{\sum_{t=1}^K \exp \left(\bm{\mu}_t^\top \bm{z}_i^v / \tau_1\right)},
\end{equation}
where $\tau_1$ is a temperature parameter. Similarly, the probability for text modality $p(k|\bm{x}_i^t)$ is defined in the same manner but uses the text feature $\bm{z}_i^t$. For notation convenience, we denote $\bm{p}_i^v = [p(1|\bm{x}_i^v), \ldots, p(K|\bm{x}_i^v)]^\top$ and $\bm{p}_i^t = [p(1|\bm{x}_i^t), \ldots, p(K|\bm{x}_i^t)]^\top$ as the prediction vectors. Then, the modality-specific networks $f_v$ and $f_t$ can be learned by minimizing the following Cross-Entropy (CE) criterion:
\begin{equation}\label{eq:ce_loss}
    \mathcal{L}_{CE} = - \frac{1}{N} \sum_{i=1}^N 
    \bm{y}_{i} \left(\log \bm{p}_i^v + \log \bm{p}_i^t\right),
\end{equation}
The solution in Eq.\eqref{eq:ce_loss} brings multimodal points with the same semantic label together. However, since $\bm{y}_i$ contains noise, minimizing Eq.\eqref{eq:ce_loss} will wrongly pull close those irrelevant samples and thus impair the cross-modal retrieval performance.

\subsection{Optimal Transport Theory}
Optimal Transport (OT) is to seek an optimal transport plan between two distributions of mass at a minimal cost. Here we briefly introduce the optimal transport theory. More details about OT can be found in \cite{peyre2019computational,villani2009optimal}. 

Let $\Delta_n = \{ \bm{x} \in \mathbb{R}_{+}^n | \sum_{i=1}^n x_i =1, \forall x_i \geq 0\}$ be the probability simplex in dimension $n$. Consider two sets of discrete data points $\bm{X} = \{x_i\}_{i=1}^m$ and $\bm{Y} = \{y_j\}_{j=1}^n$, of which the empirical probability measures are $\bm{\alpha}=\sum_{i=1}^{m}\alpha_{i}\delta(x_{i})$ and $\bm{\beta}=\sum_{j=1}^{n}\beta_{j}\delta (y_{j})$, where $\delta$ is the Dirac function. Here the weight vector $[ \alpha_{1}, \alpha_{2}, \ldots, \alpha_{m}]^\top$ and $[\beta_{1}, \beta_{2}, \ldots, \beta_{m}]^\top$ live in $\Delta_m$ and $\Delta_n$, respectively. When a meaningful cost function $c: \bm{X} \times \bm{Y} \rightarrow \mathbb{R}_{+}$ is defined, we can get the cost matrix $\bm{C} \in \mathbb{R}^{m\times n}$, where $\bm{C}_{ij} = c(x_{i}, y_{j})$. The discrete optimal transport between probability measures $\bm{\alpha}$ and $\bm{\beta}$ can be formulated as:
\begin{equation}\label{eq:ot}
\begin{aligned}
& \text{OT}(\bm{\alpha},\bm{\beta}) = \min_{\bm{T} \in \Pi(\bm{\alpha},\bm{\beta})} \langle \bm{T} ,\bm{C} \rangle_F
\\ \ \ \ & \mbox{ s.t. }  \bm{T} \mathbbm{1}_{n} = \bm{\alpha}, \bm{T}^{\top} \mathbbm{1}_{m} = \bm{\beta},
\end{aligned}
\end{equation}
where $\langle \cdot,\cdot\rangle_F$ is the Frobenius dot-product and $\mathbbm{1}_{d}$ denotes a $d$-dimensional all-one vector. $\Pi(\bm{\alpha},\bm{\beta})$ is the transport polytope that consists of all feasible transport plans. $T \in \mathbb{R}^{m\times n}$ is called the optimal transport plan, of which $T_{ij}$ denotes the amount of transport mass between $x_i$ and $y_j$ in order to obtain the overall minimal cost.

\begin{figure*}[!t]
  \centering
  \includegraphics[width=.95\textwidth]{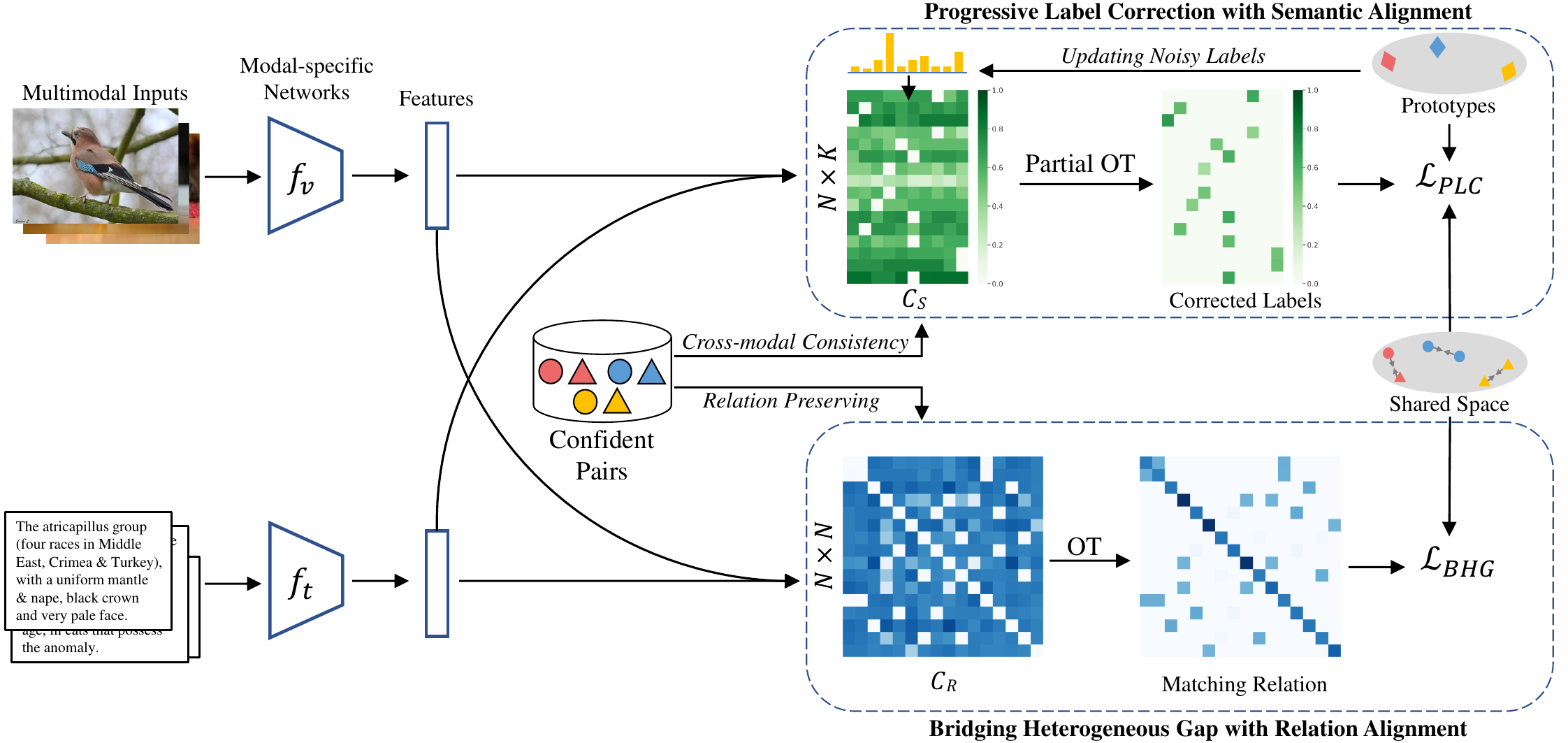}
  \caption{Illustration of UOT-RCL. Our method mainly contains two components: (1) Progressive Label Correction with Semantic Alignment that aims to mitigate the influence of noisy-labeled samples. (2) Bridging Heterogeneous Gap with Relation Alignment that aims to learn discriminative representations within the same semantics. The two components can be trained in a unified objective, facilitating robust cross-modal retrieval.}  
  \label{fig:framework}
\end{figure*}

\section{Methodology}
In this section, we present our novel OT-based unified framework for robust cross-modal learning, namely UOT-RCL. As illustrated in Fig.\ref{fig:framework}, UOT-RCL comprises two key components to mitigate the overfitting to noise (Section \ref{section:SA}) and bridge the multi-modal heterogeneous gap (Section \ref{section:RA}) respectively, where the two components are based on Optimal Transport and can be summarized in a unified framework.

\subsection{Identify Confident Cross-modal Pairs}
Our OT-based framework uses the inherent correlation among modalities to facilitate effective transport cost. Given a multi-modal sample, some modalities may provide redundant or even irrelevant information for its semantic class. Thus, we first identify confident samples that have strong cross-modal semantic consistency. Specifically, we measure the discrepancy between the bimodal predicted probabilities by quantifying its Jensen-Shannon divergence (JSD), \ie,
\begin{equation}\label{eq:JSD}
    d_i = \frac{1}{2} \text{KLD}(\bm{p}_i^v \parallel \frac{\bm{p}_i^v + \bm{p}_i^t}{2}) + \frac{1}{2} \text{KLD}(\bm{p}_i^t \parallel \frac{\bm{p}_i^v + \bm{p}_i^t}{2}),
\end{equation}
where $\text{KLD}(\cdot)$ is the KL-divergence function. Note that we warm up the training in the first few epochs with $\mathcal{L}_{CE}$ to make the networks and prototypes achieve an initial convergence.

After measuring the discrepancy, we identify confident samples with small JSD, \ie, $d_i \rightarrow 0$, indicating the bimodal data are very similar. In addition, to avoid the identified samples being concentrated on easy classes, we enforce a class balance prior by selecting samples from each class equally. Due to the presence of label noise, we take an average probability prediction over modalities to approximate the clean class. The confident set $\mathcal{G}_k$ belonging to the $k$-th class  is defined as:
\begin{equation}\label{eq:confident set}
    \mathcal{G}_k = \{(\bm{x}_i^v, \bm{x}_i^t) | d_i < \gamma_k, \forall (\bm{x}_i^v, \bm{x}_i^t)  \in \mathbbm{D}\}, k \in \{1,\cdots,K\},
\end{equation}
where $\gamma_k$ is a threshold to ensure class-balanced selecting, which is dynamically defined by the ranked JSD values. Finally, we collect the confident set over all classes as confident cross-modal pairs, \ie, $\mathcal{G} = \cup_{k=1}^K \mathcal{G}_k$.

\subsection{Progressive Label Correction with Semantic Alignment}\label{section:SA} 

\subsubsection{Cross-modal Consistent Cost Function}
To combat label noise, we view the label correction procedure as an OT problem that transports noisy labeled samples to correct semantics at a minimum cost. Denoted the corrected label matrix as $\hat{\bm{Y}}=[\hat{\bm{y}}_1,\ldots,\hat{\bm{y}}_N]^{\top} = [\hat{y}_{ij}]_{N \times K}$, the corresponding cost matrix $\bm{C}_S = [\bm{C}^S_1, \ldots, \bm{C}^S_N]^{\top} = [C^S_{ij}]_{N \times K}$ is desirable to measure the distance between $N$ samples and  $K$ semantic content. $\bm{C}_S$ can be flexibly defined in different ways; a simple yet effective option is to set $\bm{C}_S =  -\log \bm{P}$, where $\bm{P} = [\bm{p}_1,\ldots,\bm{p}_N]^{\top} = [P_{ij}]_{N \times K}$ is a proper prediction matrix, \eg, the model’s predictions \cite{peng2022optimal,wang2022solar}. However, estimating reliable predictions is challenging in our scenario. First, the incorrection in the label space imposes formidable obstacles for proper predictions. Second, multiple modalities may contribute differently to the predictions, \eg, the text is usually subjective and contains some redundant description.

To address the above issue, we propose a cross-modal consistent cost function based on the estimation of the contrastive neighbors.
Specifically, we first re-estimate the predictive reliability of a candidate sample by contrasting it against its intra-modality nearest neighbors in the feature space. Such contrastive estimation idea is based on the deep clustering \cite{alwassel2020self,zhang2021discovering} that samples with the same semantic content are aggregated in the embedding space.
For this goal, given two arbitrary representations $\bm{z}_i$ and $\bm{z}_j$ in the shared space, we measure the similarity between them by the cosine distance:
\begin{equation}\label{eq:cosine}
	d(\bm{z}_i, \bm{z}_j)= \frac{(\bm{z}_i)^{\top} \bm{z}_j}{\lVert \bm{z}_i \rVert \lVert \bm{z}_j \rVert}.
\end{equation}
Then, for the cross-modal pair $(\bm{x}_i^v, \bm{x}_i^t, \bm{y}_i)$, we contrast each modality with its top-$\mathcal{K}$ nearest neighbors based on the representation similarity to estimate the clean semantic classes, where the similarity itself is viewed as the clustering weight. To tackle the second challenge, we use the confident pairs to quantify the cross-modal consistency, which is further used to calculate an optimal blending of multi-modal predictions. Taking the visual modality for example, given $(\bm{x}_i^v, \bm{x}_i^t)$, we first use $\bm{x}_i^v$ as a query to search its intra-modal closet sample $\bm{x}_{\lozenge}^v$ from $\mathcal{G}$. Since $\bm{x}_{\lozenge}^v$ and $\bm{x}_{\lozenge}^t$ share the same semantics, the visual consistency can be measured by comparing the closet visual feature distance and the corresponding textual distance:
\begin{equation}\label{eq:cosistency}
	\mathcal{C}_i^v = \frac{d\left( f_v \left( \bm{x}_i^v \right), f_v \left( \bm{x}_{\lozenge}^v \right) \right)}{d\left( f_t \left( \bm{x}_i^t \right), f_t \left( \bm{x}_{\lozenge}^t \right) \right)}, (\bm{x}_{\lozenge}^v, \bm{x}_{\lozenge}^t) \in \mathcal{G}.
\end{equation}
Similarly, we can calculate its textual consistency $\mathcal{C}_i^t$ by using $\bm{x}_i^t$ as a query. Finally, blending the contrastive neighbors estimation with the modal consistency, we define $\bm{C}_S$ as:
\begin{equation}\label{eq:cost}
\begin{aligned}
	&\bm{C}^S_{i} = - \log  \biggl[ \mathcal{C}_i^v  \frac{1}{\mathcal{K}} \sum_{k = 1}^{\mathcal{K}} d \bigl(f_v( \bm{x}_i^v ), f_v ( \bm{x}_{k}^v ) \bigl) \bm{y_{k}} \ + 
	\\
	&  \mathcal{C}_i^t  \frac{1}{\mathcal{K}} \sum_{\hat{k} = 1}^{\mathcal{K}}  d\bigl(f_t ( \bm{x}_i^t), f_t( \bm{x}_{\hat{k}}^t ) \bigl) \bm{y}_{\hat{k}} \biggl], \bm{x}_{k}^v \in \mathcal{N}_i^v, \bm{x}_{\hat{k}}^t \in \mathcal{N}_i^t,
\end{aligned}
\end{equation}	
where $\mathcal{N}_i^v$ and $\mathcal{N}_i^t$ denotes the feature-space neighbor set of $\mathcal{K}$ nearest samples w.r.t. $\bm{x}_i^v$ and $\bm{x}_i^t$, respectively.

To enable $\bm{C}_S$ to be dynamically updated, we propose a softened and moving-average strategy to gradually update the targets:
\begin{equation}\label{eq:target_update}
    \bm{y}_i = \gamma \bm{y}_i + (1-\gamma)\bm{\upsilon},\ \ 
    \hspace{-0.1cm}\upsilon_j=\begin{cases} 
1& \hspace{-0.2cm} \text{if }j= \hspace{-0.1cm} \mathop{\arg\max}\limits_{{k\in \{1,\dots,K\}}} \bm{\mu}_k^\top (\bm{z}_i^v + \bm{z}_i^t),\\
0& \hspace{-0.2cm} \text{else}
\end{cases}
\end{equation}
where $\gamma \in [0,1]$ is the momentum parameter. The intuition is that the prototype represents the clustering center embedding for the corresponding class and thus can be used to smoothly update the training targets towards accurate ones.

\subsubsection{Progressive Label Correction as an OT Problem}
As $\bm{C}_S$ relies on the gradually updated training targets, it cannot completely mitigate the influence of noise, especially in the early stages of training. To this end, we propose to view the label correction procedure as a partial OT problem \cite{tai2021sinkhorn, chapel2020partial} that seeks only $0 \leq s \leq 1$ unit mass to be transported between samples and semantic classes, where $s$ is progressively increased over the course of training. Such progressive label correction can be formalized as the following objective:
\begin{align}
    \underset{\hat{\bm{Y}},\bm{u},\bm{w},\epsilon}{\mathrm{min}}& \ \langle \hat{\bm{Y}} ,\bm{C}_S \rangle_F  \label{eq:LP}\\
    \text{s.t.}\quad&  \hat{Y}_{ij}\geq 0, \bm{u} \succeq 0, \bm{w} \succeq 0, \epsilon \geq 0, \nonumber \\
    & \vphantom{\frac11}\hat{\bm{Y}} \ones{K} + \bm{u} = \frac{1}{N} \ones{N}, \label{eq:cons1}\\ 
    & \hat{\bm{Y}}^\top \ones{N} + \bm{w} = \bm{r}, \label{eq:cons2} \\  
    &\vphantom{\frac11}\ones{N}^\top \hat{\bm{Y}} \ones{K} = s + \epsilon. \label{eq:cons3}
\end{align}
Here the marginal probability vector $\bm{r} \in \Delta_K$ represents the expected class distribution, \eg, $\bm{r} = [\frac{1}{K}, \ldots, \frac{1}{K}]$ for a balanced classes. The row constraint Eq.\eqref{eq:cons1} indicates that the $N$ training examples are sampled uniformly. The column constraint Eq.\eqref{eq:cons2} enforces the distribution of corrected labels to match the marginal class distribution. The mass constraint Eq.\eqref{eq:cons3} restricted that the corrected labels only assign to $s$-unit most confidently samples based on transport cost. Besides, $\bm{u} \in \mathbb{R}^N$, $\bm{w} \in \mathbb{R}^K$, and $\epsilon \in \mathbb{R}$ are the optimizable variables to ensure the objective be feasible. As $\hat{\bm{Y}}$ in Eq.\eqref{eq:LP} is essentially a joint probability matrix, we re-scaled it to $N \hat{\bm{Y}}$ to represent the formal label-space matrix.

Exactly solving Eq.\eqref{eq:LP} is computationally expensive. Fortunately, it can be transformed into a standard OT problem and be solved by the efficient Sinkhorn-Knopp algorithm\cite{cuturi2013sinkhorn}. Specifically, we first collect the optimizable variables into $\bm{\tilde{Y}}$ and denote the corresponding cost matrix as $\bm{\tilde{C}_S}$, \ie,
\begin{equation*}
\bm{\tilde{Y}} \coloneqq \begin{bmatrix} \hat{\bm{Y}} & \bm{u} \\ \bm{w}^\top  & \epsilon \end{bmatrix}, \quad \bm{\tilde{C}_S} \coloneqq \begin{bmatrix} \bm{C}_S &  \zeros{N} \\ \zeros{K}^\top & 0 \end{bmatrix},
\end{equation*}
where $\zeros{d}$ denotes a $d$-dimensional all-zero vector. The row and column marginal projection measures for $\bm{\tilde{Y}}$, \ie, $\bm{\alpha}_S$ and $\bm{\beta}_S$, is defined as $[\frac{1}{N} \ones{N}^{\top}, 1-s]^{\top}$ and $[\bm{r}^{\top}, 1-s]^{\top}$ to satisfy the constraints in Eq.\eqref{eq:LP}, respectively. Based on these, computing the corrected label matrix in partial OT boils down to solving the following standard OT problem:
\begin{equation}\label{eq:semantic_align_OT}
\begin{aligned}
&\ \ \ \ \ \ \ \min_{\bm{\tilde{Y}} \in \Pi(\bm{\alpha}_S,\bm{\beta}_S)} \langle \bm{\tilde{Y}} ,\bm{\tilde{C}_S} \rangle_F \\
& \mbox{ s.t. }  \bm{\tilde{Y}} \mathbbm{1}_{k+1} = \bm{\alpha}_S, \bm{\tilde{Y}}^{\top} \mathbbm{1}_{N+1} = \bm{\beta}_S,
\end{aligned}
\end{equation}

The optimal transport plan $\bm{\tilde{Y}}$ can be solved by the efficient Sinkhorn algorithm that only contains matrix multiplication operations, and our objective $\hat{\bm{Y}}$ = $\bm{\tilde{Y}}[1:N,1:K]$. Then, we can train the modal-specific networks and prototypes robustly with corrected labels:
\begin{equation}\label{eq:corrected_loss}
    \mathcal{L}_{PLC} = - \frac{1}{N} \sum_{i=1}^N 
    N \hat{\bm{Y}}_i \left(\log \bm{p}_i^v + \log \bm{p}_i^t\right).
\end{equation}

To increase the strength of corrected samples, we increase the value of $s$ as training progresses. While there are several strategies to increase $s$ with iteration, we use a linear schedule specified by a start and end value for simplicity.

\subsection{Bridging Heterogeneous Gap with Relation Alignment}\label{section:RA} 
As mentioned, the core of cross-modal retrieval is to make correlated data from different modalities be similar in the shared space. However, the multi-modal data contains the inherent heterogeneity gap that causes inconsistent representations of various modalities. To learn discriminative representations, recent advances \cite{hu2021learning,xu2022early} resort to contrastive learning that maximizes the multi-modal mutual information at the instance level. Despite their promise, they view cross-modal data from different instances as irrelevant pairs, which hinders effective supervised CMR, \ie, learning discriminative representations within the semantic classes. 

To narrow the heterogeneity gap, we model an OT problem to infer the semantic-level matching among different modalities. In particular, we preserve the relation of each training sample to the set of confident pairs to benefit the transport plan against label noise. For the $i$-th training pair $(\bm{x}_i^v,\bm{x}_i^t)$, its visual-modal relation score to the $j$-th confident pair $(\bm{x}_{\mathcal{G}_j}^v,\bm{x}_{\mathcal{G}_j}^t)$ is defined as:
\begin{equation}\label{eq:relation score v}
    R_{i,\mathcal{G}_j}^v = \frac{\exp\bigl({-d \bigl(f_v(\bm{x}_i^v ), f_v( \bm{x}_{\mathcal{G}_j}^v) \bigl)/\tau_2}\bigl)}{\sum_{j^{\prime}=1}^{\lvert \mathcal{G} \rvert}\exp\bigl({-d\bigl(f_v(\bm{x}_i^v ), f_v( \bm{x}_{\mathcal{G}_{j^{\prime}}}^v)\bigl)/\tau_2}\bigl)}, \forall \bm{x}_{\mathcal{G}_j}^v \in \mathcal{G},
\end{equation}
where $\tau_2$ is a temperature parameter. The textual-modal relation score $R_{i,\mathcal{G}_j}^t$ can be computed analogously. We denote $\bm{R}_{i}^v  = [R_{i,\mathcal{G}_1}^v,\ldots, R_{i,\mathcal{G}_{\lvert \mathcal{G} \rvert}}^v]$ and $\bm{R}_{i}^t  = [R_{i,\mathcal{G}_1}^t,\ldots, R_{i,\mathcal{G}_{\lvert \mathcal{G} \rvert}}^t]$ to represent the relation of $\bm{x}_i^v$ and $\bm{x}_i^t$ to the confident pairs in visual and textual modalities respectively. Intuitively, if two cross-modal samples share a similar relation to the confident set in the corresponding modality, they can be considered following the same semantic information. Thus, we formalize a relation-based OT problem to seek a proper cross-modal matching $\bm{M} = [\bm{M}_1,\ldots,\bm{M}_N]^{\top} = [M_{ij}]_{N \times N}$, \ie,
\begin{equation}\label{eq:relation_align_OT}
\begin{aligned}
&\min_{\bm{M} \in \Pi(\bm{\alpha}_R,\bm{\beta}_R)} \langle \bm{M} ,\bm{C_R} \rangle_F \\
& \ \ \ \ \ \ \mbox{ s.t. } \ \ \  \bm{M} \mathbbm{1}_{N} = \bm{\alpha}_R, \bm{M}^{\top} \mathbbm{1}_{N} = \bm{\beta}_R,
\end{aligned}
\end{equation}
where $\bm{\alpha}_R$ and $\bm{\beta}_R$ are set to $\frac{1}{N} \ones{N}$ that considers the visual and textual samples follow a uniform distribution. We define the relation-based transport cost $\bm{C_R} \in \mathbb{R}^{N \times N}$ by $C^R_{ij} = \text{JSD}(\bm{R}_{i}^v,\bm{R}_{i}^t)$ based on the Jensen-Shannon divergence. 

To learn semantic-level discriminative representations, we use the solved matching relation to align cross-modal samples. Formally, we first define the inter-modal matching probability, \ie, $j$-th textual sample w.r.t. the $i$-th visual query as $p^{v2t}_{ij} = \frac{\exp({\bm{z}_i^v}^\top \bm{z}_j^t /\tau_3)}{\sum_{j^\prime =1}^N \exp({\bm{z}_i^v}^\top \bm{z}_{j^\prime}^t /\tau_3)}$, where $\tau_3$ is the temperature parameter. Symmetrically, the 
matching probability $p^{t2v}_{ij}$ is defined in a similar manner. We denote $\bm{p}^{v2t}_{i} = [p^{v2t}_{i1}, \ldots, p^{v2t}_{iN}]$ and $\bm{p}^{t2v}_{i} = [p^{t2v}_{i1}, \ldots, p^{t2v}_{iN}]$ as probability vectors. Then, we eliminate the cross-modal discrepancy by maximizing the agreement between different modalities within the same semantics, which can be formulated as the contrastive learning (InfoNCE \cite{oord2018representation}) loss:
\begin{equation}\label{eq:InfoNCE_loss}
    \mathcal{L}_{BHG} = - \frac{1}{N} \sum_{i=1}^N (\bm{M}_i^{v2t} \log \bm{p}^{v2t}_{i} + \bm{M}_i^{t2v} \log \bm{p}^{t2v}_{i} ),
\end{equation}
where $\bm{M}_i^{v2t}$ and $\bm{M}_i^{t2v}$ be the row-wise and column-wise normalized matching relation for $i$-th sample, respectively.

\subsection{The Unified Training Objective}
Combined the above analysis, the final loss function can be formulated as:
\begin{equation}\label{eq:final_loss}
    \mathcal{L} = \mathcal{L}_{PLC} + \lambda \mathcal{L}_{BHG}.
\end{equation}
By minimizing the unified loss function, the modal-specific networks can simultaneously mitigate the impact of noisy-labeled samples and bridge the heterogeneous gap, and thus achieve robust cross-modal retrieval.

\section{Experiment}
In this section, we conduct comprehensive experiments to verify the effectiveness of our proposed method.
\subsection{Datasets and Evaluation Protocol} 
\subsubsection{Datasets.}Without loss of generality, we evaluate our method on three widely-used cross-modal retrieval datasets. The details are introduced as follows:
\begin{itemize}
  \item \textbf{Wikipedia} \cite{rasiwasia2010new} collects 2,866 corresponding image-text pairs that belong to 10 classes. Following the split in \cite{feng2014cross}, we use 2,173 pairs for training, 231 pairs for validation, and 462 pairs for testing.
  \item \textbf{NUS-WIDE} \cite{chua2009nus} is a large-scale multi-label dataset containing 269,648 image-text pairs distributed over 81 classes. In the experiments, we employ the same subsets following \cite{hu2021learning}, wherein each pair only belongs to one of ten classes. Specifically, the subsets contain 42,941, 5,000, and 23,661 pairs for training, validation, and testing, respectively.
  \item \textbf{XMediaNet} \cite{peng2018modality} is a multimodal dataset consisting of five media types, \ie{image, text, video, audio, and 3D model}. Following \cite{hu2021learning}, we select image and text modalities with 200 categories to conduct experiments. Specifically, we randomly divide the dataset into three subsets, \ie{32,000 pairs for training, 4,000 pairs for validation, and 4,000 pairs for testing.}
\end{itemize}

\subsubsection{Evaluation Protocol.} We evaluate the retrieval performance with the Mean Average Precision (mAP). In a nutshell, mAP measures the accuracy scores of retrieval results, which computes the mean value of Average Precision (AP) for each query, \ie,
\begin{equation}\label{eq:mAP}
    mAP = \frac{1}{K} \sum_{k=1}^K \frac{{TP}_k}{{FP}_k + {TP}_k},
\end{equation}
where ${FP}_k$ and ${TP}_k$ denotes true positives and false positives for the $k$-th class, respectively. In our experiments, we take images and texts as queries to retrieve the relevant texts and image samples, respectively. For all methods, we choose the best checkpoint on the validation set and report the corresponding performance on the test set. Following \cite{hu2021learning}, we set the label noise to be symmetric with noise ratios 0.2, 0.4, 0.6, and 0.8 to evaluate the retrieval performance of the methods comprehensively.

\subsection{Implementation Details}
As a general framework, our UOT-RCL can be readily applied to various CMR methods to improve their robustness. For a fair comparison, we employ the same network backbones with MRL \cite{hu2021learning}, \ie the VGG-19 \cite{simonyan2014very} pre-trained on ImageNet \cite{deng2009imagenet} as the backbone for images, the Doc2Vec \cite{lau2016empirical} pre-trained on Wikipedia as the backbone for text, and a three fully-connected layers stacked after the backbones to learn common embeddings among multiple modalities. The dimension of common space $L$ is set as 512. For all datasets, the training processes contain 100 epochs, and the warm-up training has 2 epochs. We use Adam as our optimizer, and the learning rate is set to $10^{-4}$ for all experiments. The batch size is set as 100, 500, and 100 for Wikipedia, NUS-WIDE, and XMediaNet, respectively. The temperature parameters $\tau_1$, $\tau_2$, and $\tau_3$ are set as 1. The momentum parameter $\gamma$ is set as 0.99, and we set the weight factor $\lambda$ to 0.4. The number of neighbors is approximately 5\textperthousand \ of training data, \ie we set $\mathcal{K}$ as 10, 200, and 150 for Wikipedia, NUS-WIDE, and XMediaNet, respectively. The number of samples belonging to each class in the confident set $\mathcal{G}$ is 5 for all experiments. For our progressive label correction process, the mass parameter $s$ is linearly increased from 0.2 to 0.8 over the course of training in all experiments.

\subsection{Comparison with the State-of-the-Art}
We compare UOT-RCL with 15 state-of-the-art CMR methods, including four alignment-wise methods (\ie MCCA \cite{rupnik2010multi}, PLS \cite{sharma2011bypassing}, DCCA \cite{andrew2013deep}, and DCCAE \cite{wang2015deep}), and ten label-wise methods (\ie MvDA \cite{kan2015multi}, GSS-SL \cite{zhang2017generalized}, ACMR \cite{wang2017adversarial}, deep-SM \cite{wei2016cross}, FGCrossNet \cite{he2019new}, SDML \cite{hu2019scalable}, DSCMR \cite{zhen2019deep}, SMLN \cite{hu2020semi}, MRL \cite{hu2021learning}, and ELRCMR \cite{xu2022early}). Note that MRL and ELRCMR are designed to tackle CMR with noisy labels. Table~\ref{tab:table_wiki} and Table~\ref{tab:table_nus_xm} present the retrieval mAP scores under a range of noisy label ratios on three datasets, respectively. As shown in these results, we can obtain the following observations:
\begin{itemize}
  \item Training with noisy labels can remarkably harm the performance of supervised CMR models. As the ratio of label noise increases, the mAP scores of these methods will decrease rapidly.
  \item Even if the labels are noisy, unsupervised methods are usually inferior to supervised counterparts, which indicates that unsupervised methods are poor at learning semantic-level discriminative representations. However, unsupervised methods work better in the case of extreme noisy labels, \ie, 80\% ratio.
  \item Our method outperforms all existing state-of-the-art methods on all datasets with different noise settings, which shows the superior robustness of UOT-RCL against noisy labels. Moreover, when the noise ratio is high, the improvement of UOT-RCL is more evident, which can be observed in Wikipedia Table~\ref{tab:table_wiki}.
  \item An increased number of classes makes learning with noisy labels more challenging. As shown in Table~\ref{tab:table_nus_xm}, even some robust methods, \eg, MRL, demonstrate a significant performance drop on XMediaNet with high noise. By contrast, our UOT-RCL consistently achieves superior results, \eg, we surpass the best baseline under different noisy settings by 7.4\%, 5.7\%, 5.9\%, and 4.3\%, respectively. 

\end{itemize}

\begin{table*}[t]
    \setlength\tabcolsep{5pt}
    \centering
    \caption{Retrieval performance comparison on NUS-WIDE and XMediaNet datasets under the symmetric noise rates of 20\%, 40\%, 60\%, and 80\%. Bold indicates superior results.}
    \begin{tabular}{l|cccc|cccc|cccc|cccc}
    \hline
          & \multicolumn{8}{c|}{NUS-WIDE}                                 & \multicolumn{8}{c}{XMediaNet} \\
\cline{2-17}    Method & \multicolumn{4}{c|}{Image-to-Text} & \multicolumn{4}{c|}{Text-to-Image} & \multicolumn{4}{c|}{Image-to-Text} & \multicolumn{4}{c}{Text-to-Image} \\
          & 0.2   & 0.4   & 0.6   & 0.8   & 0.2   & 0.4   & 0.6   & 0.8   & 0.2   & 0.4   & 0.6   & 0.8   & 0.2   & 0.4   & 0.6   & 0.8  \\
    \hline
    MCCA  & 0.523  & 0.523  & 0.523  & 0.523  & 0.539  & 0.539  & 0.539  & 0.539  & 0.233  & 0.233  & 0.233  & 0.233  & 0.249  & 0.249  & 0.249  & 0.249  \\
    PLS   & 0.498  & 0.498  & 0.498  & 0.498  & 0.517  & 0.517  & 0.517  & 0.517  & 0.276  & 0.276  & 0.276  & 0.276  & 0.266  & 0.266  & 0.266  & 0.266  \\
    DCCA  & 0.527  & 0.527  & 0.527  & 0.527  & 0.537  & 0.537  & 0.537  & 0.537  & 0.152  & 0.152  & 0.152  & 0.152  & 0.162  & 0.162  & 0.162  & 0.162  \\
    DCCAE & 0.529  & 0.529  & 0.529  & 0.529  & 0.538  & 0.538  & 0.538  & 0.538  & 0.149  & 0.149  & 0.149  & 0.149  & 0.159  & 0.159  & 0.159  & 0.159  \\
    \hline
    GMA   & 0.545  & 0.515  & 0.488  & 0.469  & 0.547  & 0.517  & 0.491  & 0.475  & 0.400  & 0.380  & 0.344  & 0.276  & 0.376  & 0.364  & 0.336  & 0.277  \\
    MvDA  & 0.590  & 0.551  & 0.568  & 0.471  & 0.609  & 0.585  & 0.596  & 0.498  & 0.329  & 0.318  & 0.301  & 0.256  & 0.324  & 0.314  & 0.296  & 0.254  \\
    GSS-SL & 0.639  & 0.639  & 0.631  & 0.567  & 0.659  & 0.658  & 0.650  & 0.592  & 0.431  & 0.381  & 0.256  & 0.044  & 0.417  & 0.361  & 0.221  & 0.031  \\
    ACMR  & 0.530  & 0.433  & 0.318  & 0.269  & 0.547  & 0.476  & 0.304  & 0.241  & 0.181  & 0.069  & 0.018  & 0.010  & 0.191  & 0.043  & 0.012  & 0.009  \\
    deep-SM & 0.693  & 0.680  & 0.673  & 0.628  & 0.690  & 0.681  & 0.669  & 0.629  & 0.557  & 0.314  & 0.276  & 0.062  & 0.495  & 0.344  & 0.021  & 0.014  \\
    FGCrossNet & 0.661  & 0.641  & 0.638  & 0.594  & 0.669  & 0.669  & 0.636  & 0.596  & 0.372  & 0.280  & 0.147  & 0.053  & 0.375  & 0.281  & 0.160  & 0.052  \\
    SDML  & 0.694  & 0.677  & 0.633  & 0.389  & 0.693  & 0.681  & 0.644  & 0.416  & 0.534  & 0.420  & 0.216  & 0.009  & 0.563  & 0.445  & 0.237  & 0.011  \\
    DSCMR & 0.665  & 0.661  & 0.653  & 0.509  & 0.667  & 0.665  & 0.655  & 0.505  & 0.461  & 0.224  & 0.040  & 0.008  & 0.477  & 0.224  & 0.028  & 0.010  \\
    SMLN  & 0.676  & 0.651  & 0.646  & 0.525  & 0.685  & 0.650  & 0.639  & 0.520  & 0.520  & 0.445  & 0.070  & 0.070  & 0.514  & 0.300  & 0.303  & 0.226  \\
    MRL   & 0.696  & 0.690  & 0.686  & 0.669  & 0.697  & 0.695  & 0.688  & 0.673  & 0.523  & 0.506  & 0.393  & 0.311  & 0.530  & 0.520  & 0.415  & 0.325  \\
    ELRCMR & 0.694  & 0.692  & 0.684  & 0.671  & 0.694  & 0.693  & 0.689  & 0.673  & 0.525  & 0.523  & 0.502  & 0.455  & 0.535  & 0.532  & 0.516  & 0.482 \\
    UOT-RCL & \textbf{0.706} & \textbf{0.706} & \textbf{0.690} & \textbf{0.683} & \textbf{0.705} & \textbf{0.700} & \textbf{0.695} & \textbf{0.677} & \textbf{0.558} & \textbf{0.550} & \textbf{0.532} & \textbf{0.486} & \textbf{0.576} & \textbf{0.562} & \textbf{0.545} & \textbf{0.494}  \\
    \hline
    \end{tabular}%
  \label{tab:table_nus_xm}%
\end{table*}%
\begin{table}[h]
  \renewcommand\arraystretch{1}
  \setlength\tabcolsep{2pt}
  \centering

    \caption{Retrieval performance comparison on Wikipedia dataset under the symmetric noise rates of 20\%, 40\%, 60\%, and 80\%. Bold indicates superior results.}
    \resizebox{\linewidth}{!}{
    \begin{tabular}{l|cccc|cccc}
    \hline
          & \multicolumn{8}{c}{Wikipedia} \\
\cline{2-9}    Method & \multicolumn{4}{c|}{Image-to-Text} & \multicolumn{4}{c}{Text-to-Image} \\
          & 0.2   & 0.4   & 0.6   & 0.8   & 0.2   & 0.4   & 0.6   & 0.8  \\
    \hline
    MCCA  & 0.202  & 0.202  & 0.202  & 0.202  & 0.189  & 0.189  & 0.189  & 0.189  \\
    PLS   & 0.377  & 0.377  & 0.337  & 0.337  & 0.320  & 0.320  & 0.320  & 0.320  \\
    DCCA  & 0.281  & 0.281  & 0.281  & 0.281  & 0.260  & 0.260  & 0.260  & 0.260  \\
    DCCAE & 0.308  & 0.308  & 0.308  & 0.308  & 0.286  & 0.286  & 0.286  & 0.286  \\
    \hline
    GMA   & 0.200  & 0.178  & 0.153  & 0.139  & 0.189  & 0.160  & 0.141  & 0.136  \\
    MvDA  & 0.379  & 0.285  & 0.217  & 0.144  & 0.350  & 0.270  & 0.207  & 0.142  \\
    GSS-SL & 0.444  & 0.390  & 0.309  & 0.174  & 0.398  & 0.353  & 0.287  & 0.169  \\
    ACMR  & 0.276  & 0.231  & 0.198  & 0.135  & 0.285  & 0.194  & 0.183  & 0.138  \\
    deep-SM & 0.441  & 0.387  & 0.293  & 0.178  & 0.392  & 0.364  & 0.248  & 0.177  \\
    FGCrossNet & 0.403  & 0.322  & 0.233  & 0.156  & 0.358  & 0.284  & 0.205  & 0.147  \\
    SDML  & 0.464  & 0.406  & 0.299  & 0.170  & 0.448  & 0.398  & 0.311  & 0.184  \\
    DSCMR & 0.426  & 0.331  & 0.226  & 0.142  & 0.390  & 0.300  & 0.212  & 0.140  \\
    SMLN  & 0.449  & 0.365  & 0.275  & 0.251  & 0.403  & 0.319  & 0.246  & 0.237  \\
    MRL   & 0.514  & 0.491  & 0.464  & 0.435  & 0.461  & 0.453  & 0.421  & 0.400  \\
    ELRCMR & 0.515  & 0.503  & 0.496  & 0.447  & 0.460  & 0.456  & 0.447  & 0.416  \\
    UOT-RCL & \textbf{0.522} & \textbf{0.520} & \textbf{0.503} & \textbf{0.473} & \textbf{0.469} & \textbf{0.464} & \textbf{0.454} & \textbf{0.427} \\
    \hline
    \end{tabular}%
    }
  \label{tab:table_wiki}%
\end{table}%

\begin{table*}[t]
  \centering
    \caption{Retrieval performance comparison with CLIP on XMediaNet. Bold indicates superior results.}
    \begin{tabular}{l|cccc|cccc}
    \hline
    \multirow{2}[4]{*}{Method} & \multicolumn{8}{c}{XMediaNet} \\
\cline{2-9}          & \multicolumn{4}{c|}{Image-to-Text} & \multicolumn{4}{c}{Text-to-Image} \\
          & 0.2   & 0.4   & 0.6   & 0.8   & 0.2   & 0.4   & 0.6   & 0.8  \\
    \hline
    CLIP (RN50) Zero-shot & 0.333  & 0.333  & 0.333  & 0.333  & 0.320  & 0.320  & 0.320  & 0.320  \\
    CLIP (ViT-B/16)  Zero-shot & 0.358  & 0.358  & 0.358  & 0.358  & 0.370  & 0.370  & 0.370  & 0.370  \\
    CLIP (ViT-L/14) Zero-shot & 0.377  & 0.377  & 0.377  & 0.377  & 0.403  & 0.403  & 0.403  & 0.403  \\
    \hline
    CLIP (RN50) Unsupervised fine-tune & 0.655  & 0.655  & 0.655  & 0.655  & 0.660  & 0.660  & 0.660  & 0.660  \\
    CLIP (ViT-B/16) Unsupervised fine-tune & 0.738  & 0.738  & 0.738  & 0.738  & 0.738  & 0.738  & 0.738  & 0.738  \\
    \hline
    CLIP (RN50) Supervised fine-tune & 0.674  & 0.599  & 0.481  & 0.250  & 0.681  & 0.614  & 0.520  & 0.293  \\
    CLIP (ViT-B/16)  Supervised fine-tune & 0.750  & 0.664  & 0.538  & 0.156  & 0.758  & 0.677  & 0.545  & 0.164  \\
    \hline
    CLIP (ResNet50) + UOT-RCL & 0.798  & 0.793  & 0.784  & 0.769  & 0.792  & 0.791  & 0.779  & 0.768  \\
    CLIP (ViT-B/16) + UOT-RCL & \textbf{0.825} & \textbf{0.820} & \textbf{0.815} & \textbf{0.801} & \textbf{0.819} & \textbf{0.819} & \textbf{0.816} & \textbf{0.799} \\
    \hline
    \end{tabular}%
  \label{tab:table_clip}%
\end{table*}%

\subsection{Comparison to Large Pre-trained Model}
In this section, we perform comparison to the large pre-trained vision-language model, \ie, CLIP \cite{radford2021learning}, which is trained over 400 million image-text pairs and shows powerful performance on transferring downstream tasks. Such a comparison is helpful in understanding how to adapt the large pre-trained model to downstream vision-language retrieval tasks containing noisy supervision. In the experiments, we compare CLIP on the XMediaNet dataset under the following four settings, \ie, zero-shot, unsupervised fine-tune, supervised fine-tune, and combined with our UOT-RCL framework. Specifically, the first setting directly uses trained CLIP models to conduct inference on XMediaNet, the second one uses contrastive learning to align cross-modal representations belonging to the same category from the same pair, the third one uses contrastive learning to align cross-modal representations belonging to the same category, and the last one uses our UOT-RCL framework based on trained CLIP models. All models are trained for 16 epochs with a batch size of 128, and the optimizer settings for fine-tuning follow the CLIP-FT code \footnote{\url{https://github.com/Zasder3/train-CLIP-FT}}. As shown in Table~\ref{tab:table_clip}, we could observe the following conclusions: 
\begin{itemize}
  \item Although CLIP utilizes massive pairs for pre-training, it focuses on aligning instance-level cross-modal representations and shows noteworthy performance degradation in retrieving category-level cross-modal samples.
  \item Despite the presence of noisy labels, the supervised fine-tune variants are still superior to unsupervised ones in a low noise ratio, which evidences the importance of supervision in learning semantic category discrimination.
  \item The capacity of the CLIP models affects the robustness of supervised variants to noisy labels. Larger models are more prone to overfit the noise in a high noise ratio.
  \item Applying CLIP to our framework can significantly improve its category-level cross-modal retrieval performance, \ie, UOT-RCL improves ViT-B/16 based CLIP by at least 8.1\%, 8.1\%, 7.7\%, and 6.1\% in incremental four noisy settings.

\end{itemize}

\begin{figure*}[htbp]
\setlength {\belowcaptionskip} {-0.19cm}
\centering
\subfloat[20\% Noise Ratio]{\includegraphics[width=0.25\textwidth]{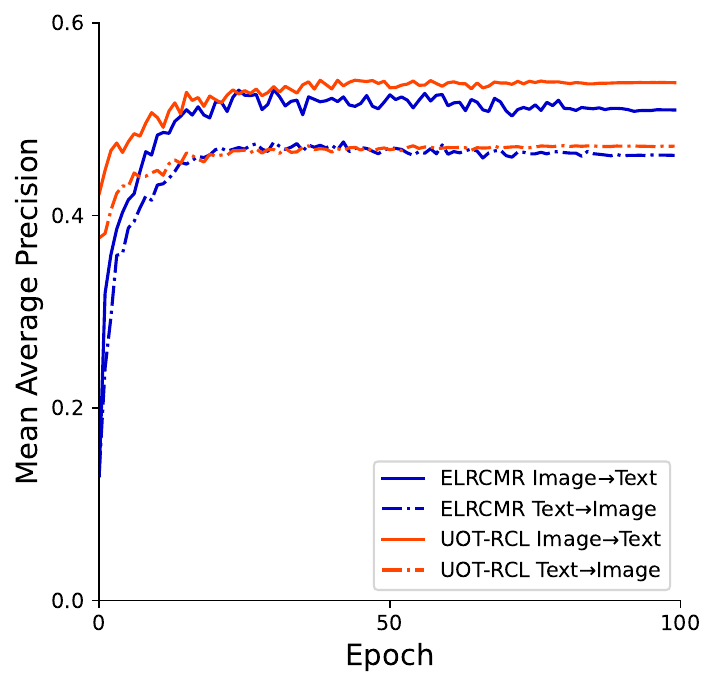}%
\label{process_2}}
\subfloat[40\% Noise Ratio]{\includegraphics[width=0.25\textwidth]{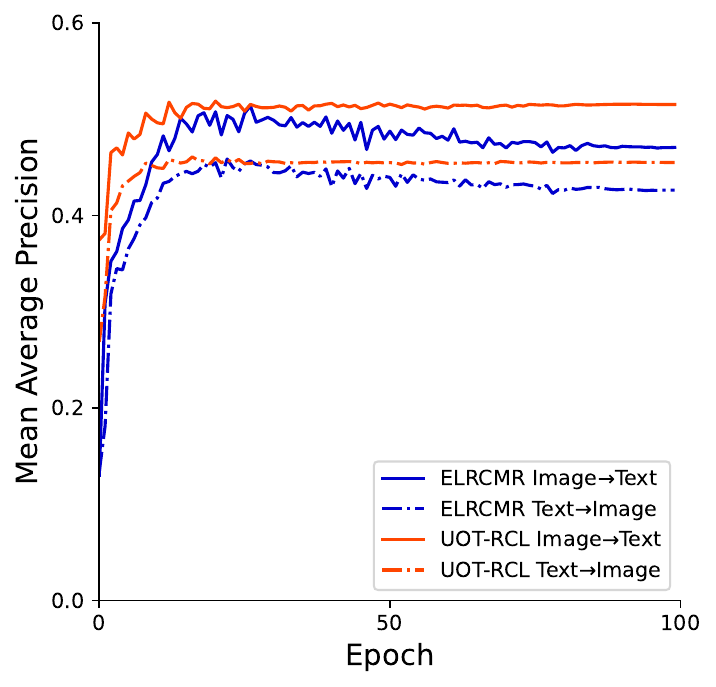}%
\label{process_4}}
\subfloat[60\% Noise Ratio]{\includegraphics[width=0.25\textwidth]{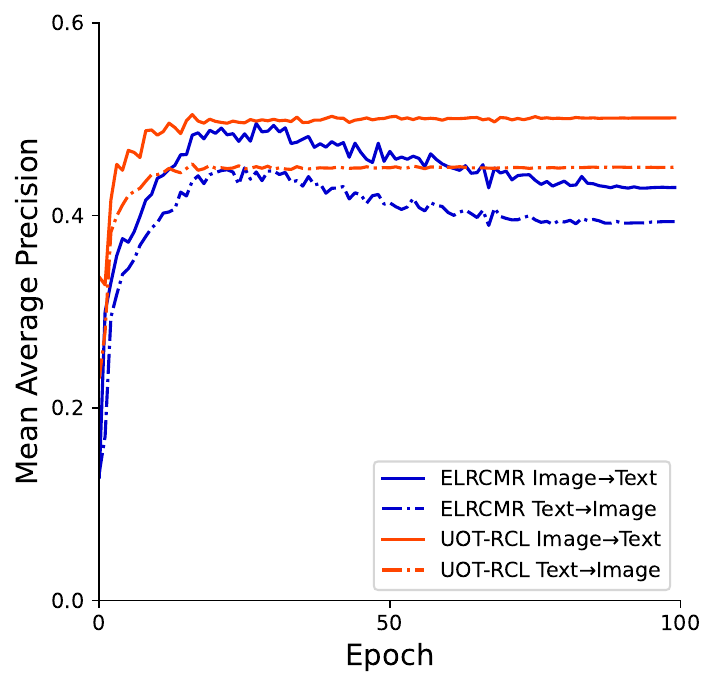}%
\label{process_6}}
\subfloat[80\% Noise Ratio]{\includegraphics[width=0.25\textwidth]{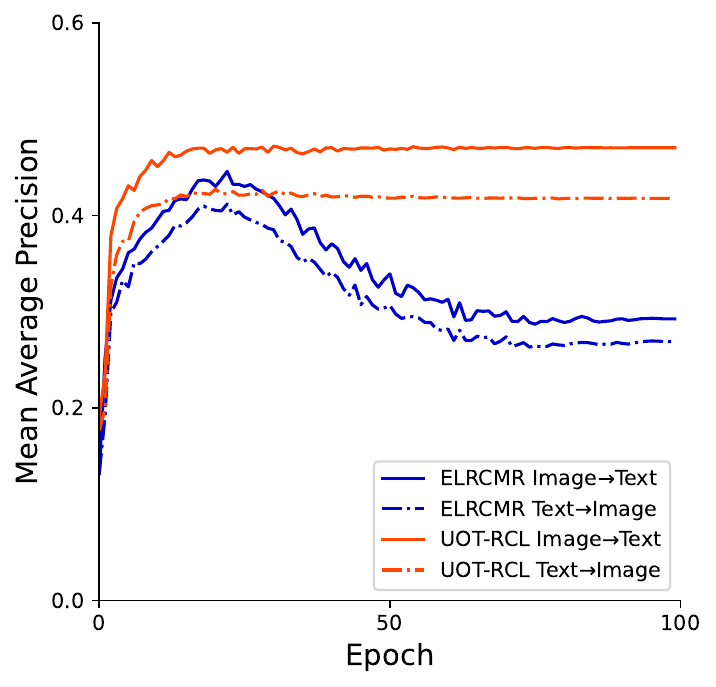}%
\label{process_8}}
\caption{Cross-modal retrieval performance of the proposed method versus ELRCMR in terms of mAP scores on the validation set of Wikipedia dataset under different noisy ratios.}
\label{training_process}
\end{figure*}

\subsection{Robustness Analysis}
To visually investigate the robustness of the proposed method,  we compare the mAP scores of our UOT-RCL and the ELRCMR method on the validation set of Wikipedia under different noise ratios. As shown in Fig.\ref{training_process}, one could see that our UOT-RCL surpasses the ELRCMR method in both performance and stability in all noisy settings, which indicates that our method has improved the robustness of the CMR model against noisy labels. Although ELRCMR adopts early learning regularization to prevent memorizing noisy labels, it still tends to overfit with training processes, especially in the high noisy settings, \ie, Fig.\ref{process_6} and Fig.\ref{process_8}. Thanks to our UOT-RCL, which generates corrected labels and narrows the heterogeneous gap, one could greatly improve the robustness of the model to resist extreme noise.

\begin{table}[H]
  \centering
  \caption{Ablation studies on Wikipedia with 0.2, 0.4, 0.6, and 0.8 noise ratios. The best results are in bold.}
    \begin{tabular}{l|cccc}
    \hline
    \multirow{2}{*}{Method} & \multicolumn{4}{c}{Image-to-Text} \\
          & 0.2   & 0.4   & 0.6   & 0.8  \\
    \hline
    UOT-RCL  & \textbf{0.522} & \textbf{0.520} & \textbf{0.503} & \textbf{0.473} \\
    UOT-RCL w/o $\mathcal{L}_{PLC}$ & 0.476  & 0.464  & 0.451  & 0.400  \\
    UOT-RCL w/o $\mathcal{L}_{BHG}$ & 0.504  & 0.499  & 0.474  & 0.399  \\
    UOT-RCL w/o confident set & 0.521  & 0.519  & 0.503  & 0.466  \\
    \hline
          & \multicolumn{4}{c}{Text-to-Image} \\
    UOT-RCL  & \textbf{0.469} & \textbf{0.464} & \textbf{0.454} & \textbf{0.427} \\
    UOT-RCL w/o $\mathcal{L}_{PLC}$ & 0.434  & 0.423  & 0.410  & 0.366  \\
    UOT-RCL w/o $\mathcal{L}_{BHG}$ & 0.453  & 0.447  & 0.420  & 0.363  \\
    UOT-RCL w/o confident set & 0.465  & 0.461  & 0.450  & 0.425  \\
    \hline
    \end{tabular}%
  \label{tab:table_ablation}%
\end{table}%

\subsection{Ablation Study}
To study the influence of specific components in our method, we carry out the ablation study on the Wikipedia dataset under different noise ratios. Specifically, we ablate the contributions of three key components in our UOT-RCL, \ie, $\mathcal{L}_{PLC}$, $\mathcal{L}_{BHG}$, and the confident set. Note that we select random samples into $\mathcal{G}$ to explore the influence of the confident set. For a fair comparison, all the compared methods are trained with the same settings as our UOT-RCL. As shown in Table~\ref{tab:table_ablation}, one could see that the full UOT-RCL achieves the best performance under different noise ratios, showing that all three components are important to improve the robustness against noisy labels. We could also see that the label correction process, \ie, $\mathcal{L}_{PLC}$, has the most important contribution to combat noise. Besides, the contribution of $\mathcal{L}_{BHG}$ increases with the rise in the noise ratios, which indicates that noisy labels can confuse the discriminative connections among different modalities and widen the heterogeneous gap.

\begin{figure*}[htb]
\setlength {\belowcaptionskip} {-0.19cm}
\centering
\subfloat[20\% Noise Ratio I2T]{\includegraphics[width=0.25\textwidth]{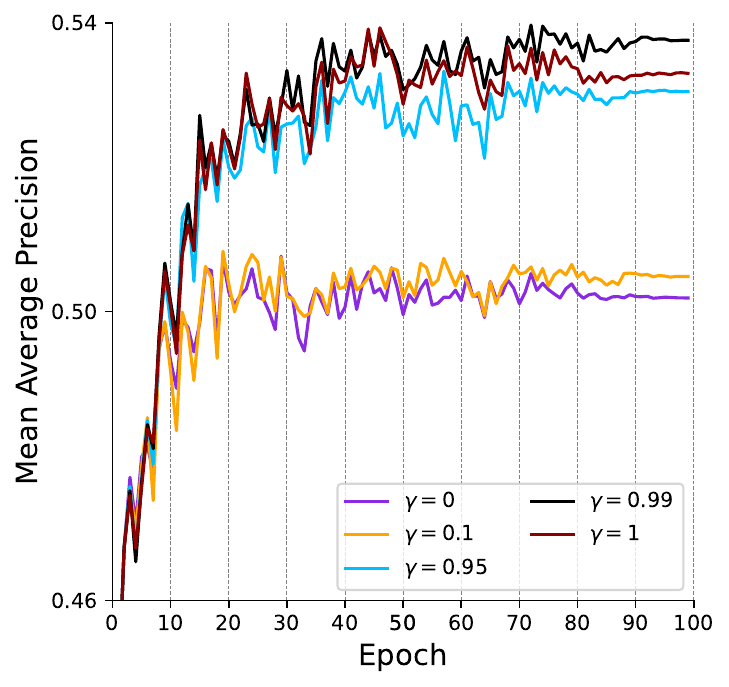}%
\label{momentum_n2_i2t}}
\subfloat[20\% Noise Ratio T2I]{\includegraphics[width=0.25\textwidth]{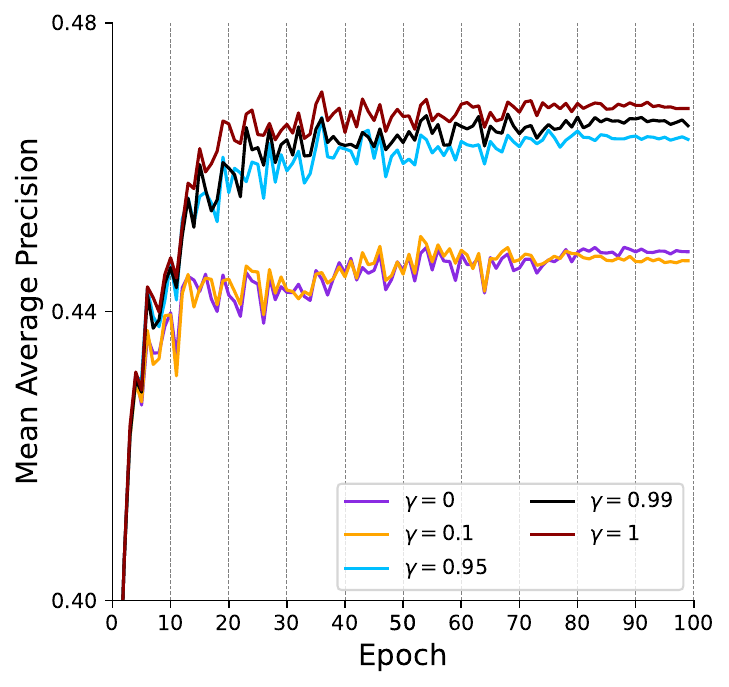}%
\label{momentum_n2_t2i}}
\subfloat[80\% Noise Ratio I2T]{\includegraphics[width=0.25\textwidth]{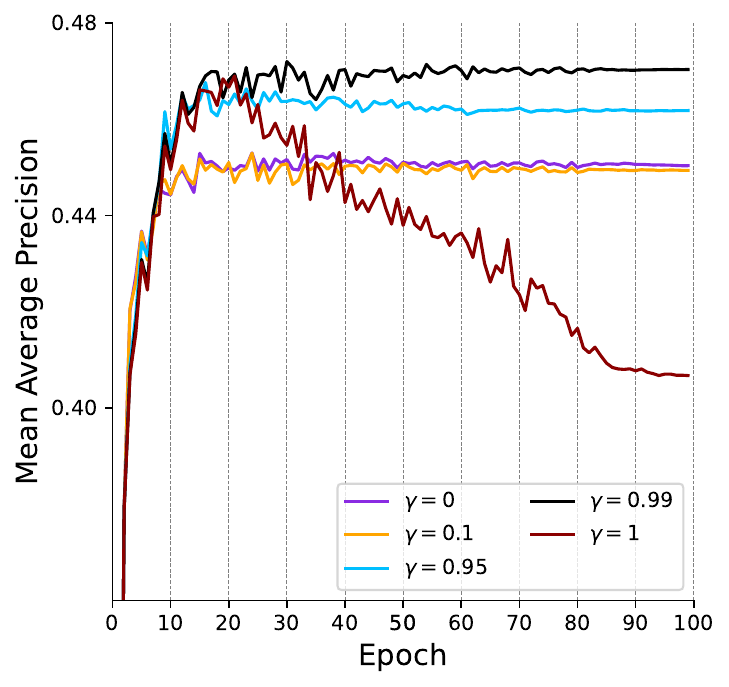}%
\label{momentum_n8_i2t}}
\subfloat[80\% Noise Ratio T2I]{\includegraphics[width=0.25\textwidth]{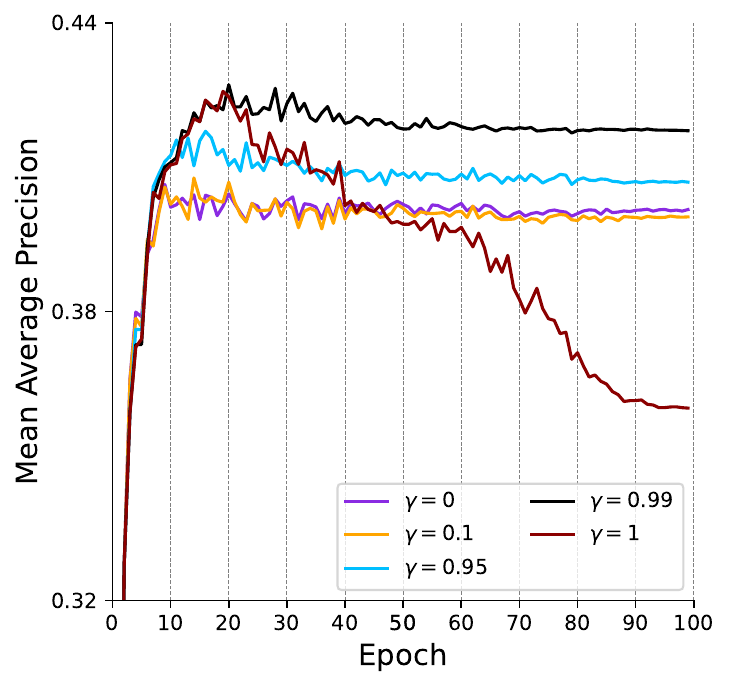}%
\label{momentum_n8_t2i}}
\caption{Cross-modal retrieval performance of our UOT-RCL in terms of mAP scores on the Wikipedia validation set with different values of $\gamma$. (a) The image-to-text performance under 20\% noise ratio. (b) The text-to-image performance under 20\% noise ratio. (c) The image-to-text performance under 80\% noise ratio. (d) The text-to-image performance under 80\% noise ratio. We set $\gamma$ to 0, 0.1, 0.95, 0.99, and 1 to study the influence on retrieval results.}
\label{momentum}
\end{figure*}

\begin{figure*}[htbp]
\centering
\subfloat[]{\includegraphics[width=0.329\textwidth]{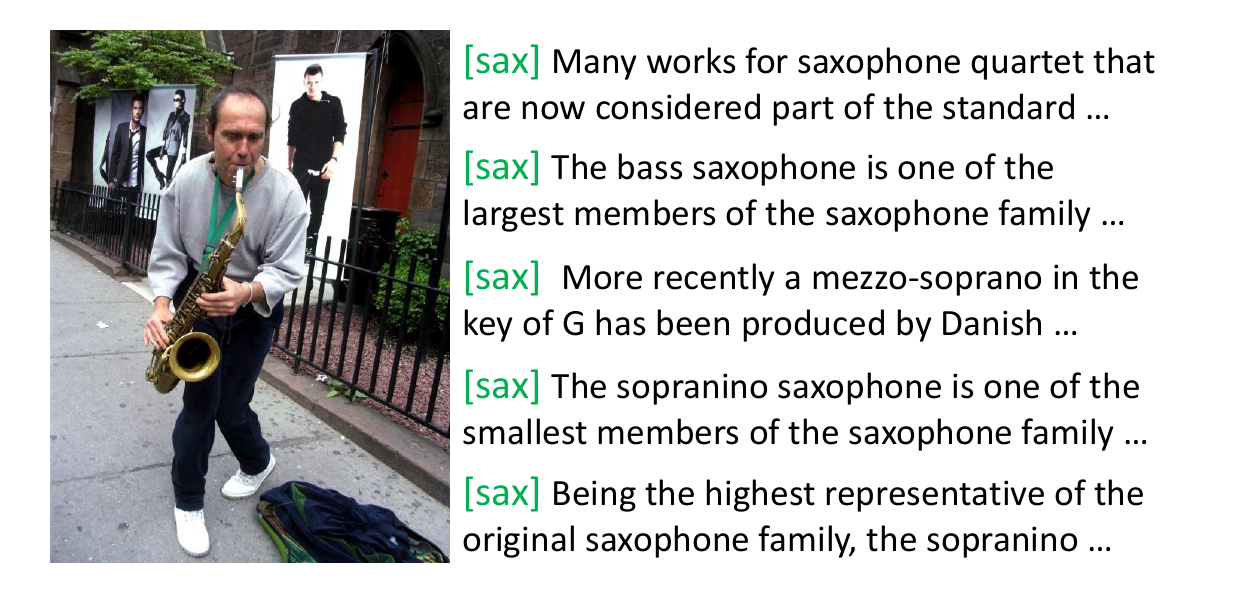}%
\label{i2t_1}}
\subfloat[]{\includegraphics[width=0.329\textwidth]{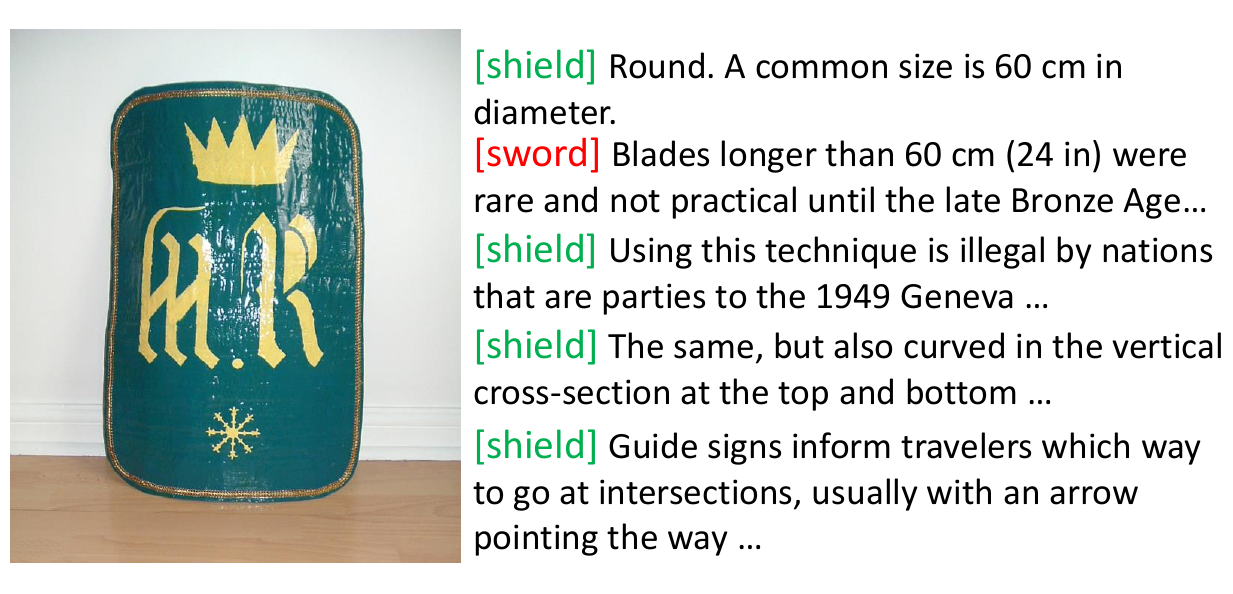}%
\label{i2t_2}}\hspace{1mm}
\subfloat[]{\includegraphics[width=0.329\textwidth]{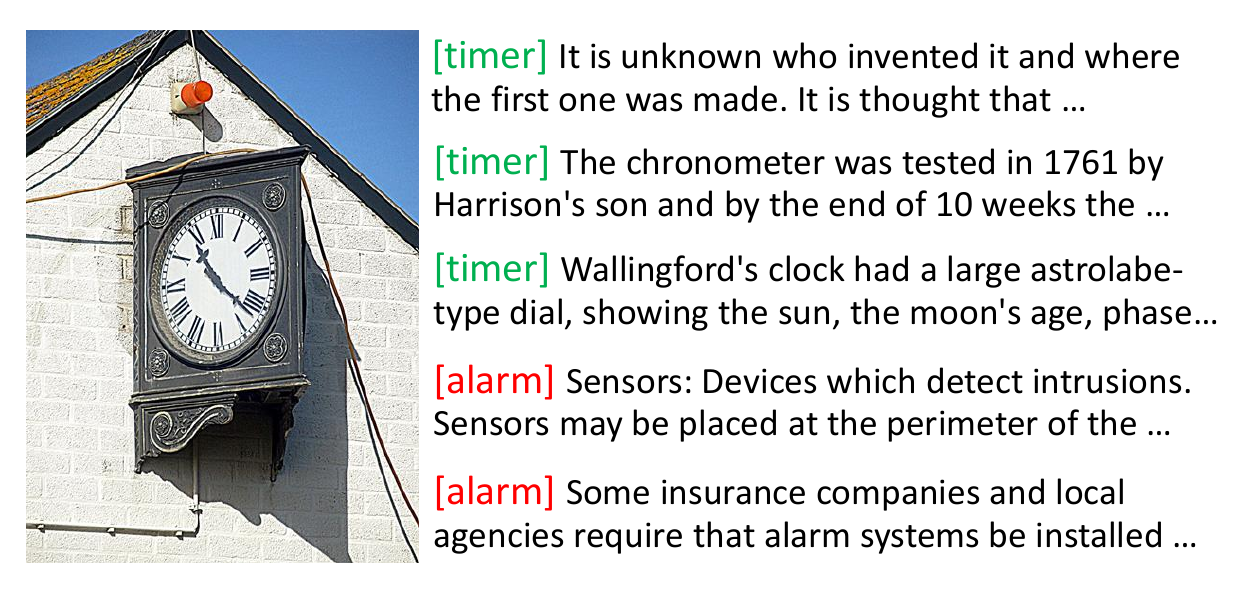}%
\label{i2t_3}}

\vspace{-.35cm}
\subfloat[]{\includegraphics[width=0.5\textwidth, trim=1mm 0mm 1mm 0mm, clip=true]{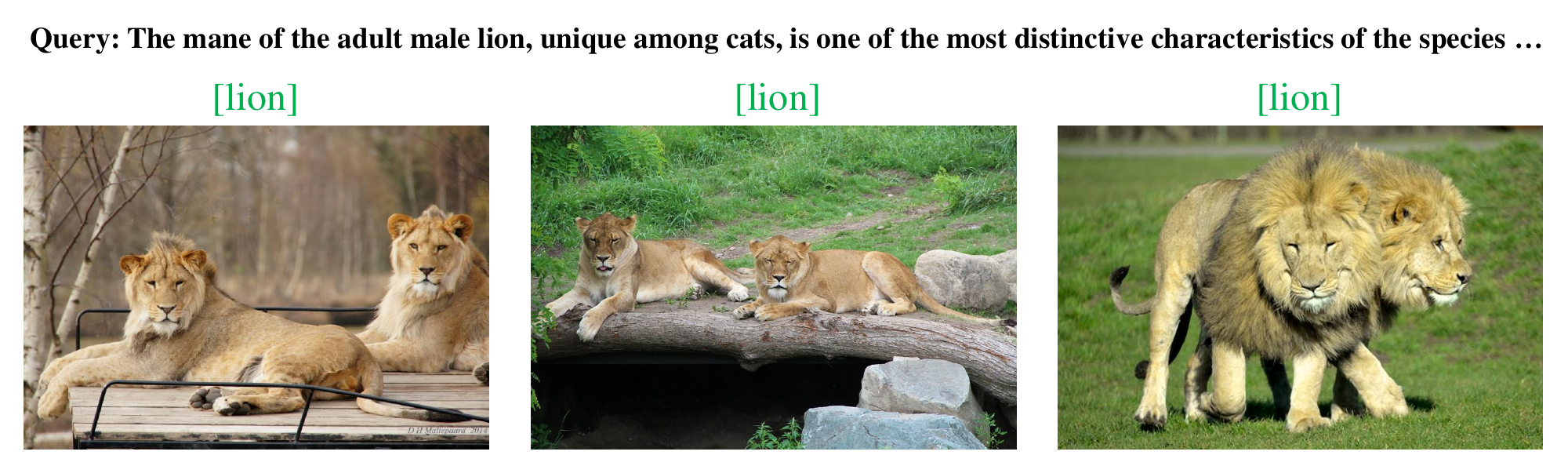}%
\label{t2i_1}}
\subfloat[]{\includegraphics[width=0.5\textwidth,trim=1mm 0mm 1mm 1.5mm, clip=true]{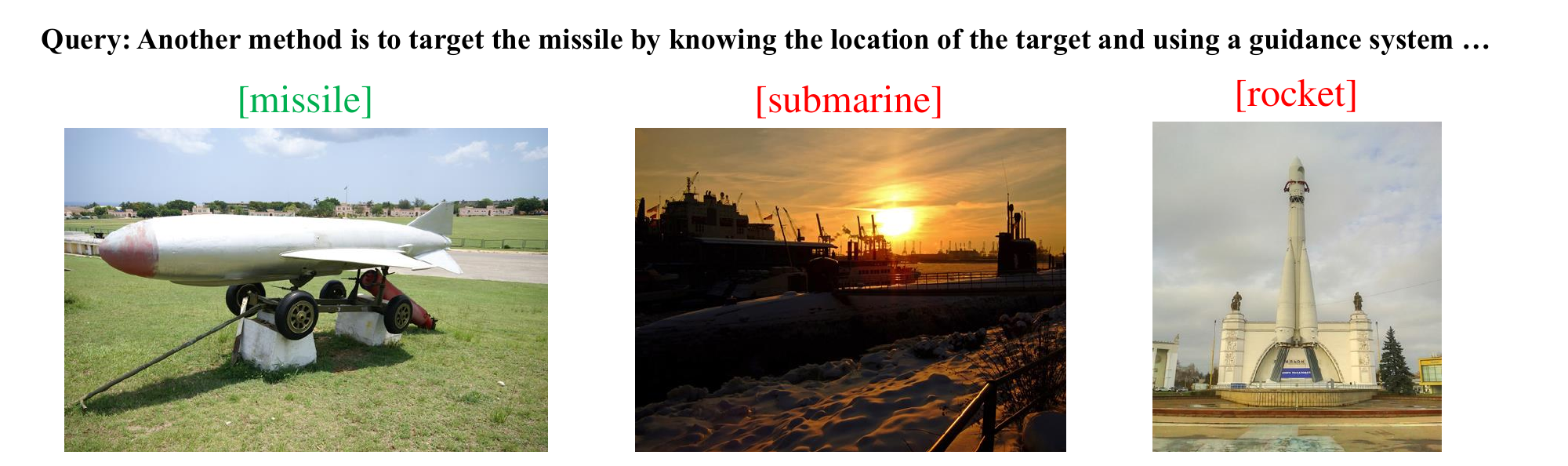}%
\label{t2i_2}}
\caption{Some retrieval cases on XMediaNet under 40\% noise ratio. For each image query, \ie (a)-(c), we show the top-5 ranked texts. For each text query, \ie (d)-(e), we show the top-3 ranked images. We mark the corresponding label of the correct retrieval result in green and otherwise in red.}
\label{case}
\end{figure*}

\begin{figure}[h]
\centering  
\includegraphics[width=0.85\columnwidth]{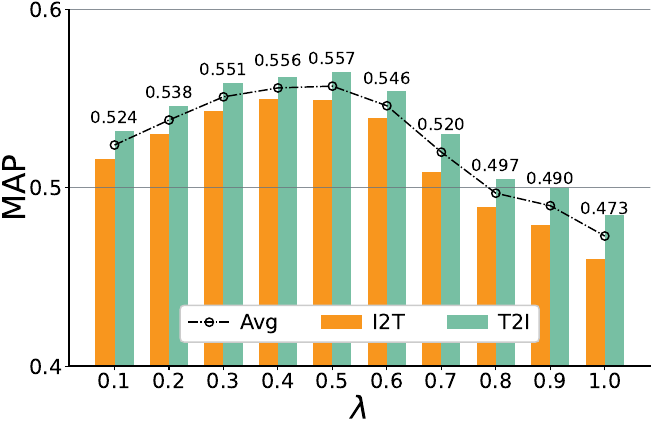}
\caption{Cross-modal retrieval performance of our UOT-RCL in terms of mAP scores versus different values of the weight factor $\lambda$ on the testing set of the XMediaNet. The weight factor is incremental from 0.1 to 1 in intervals of 0.1. The noise ratio is 0.4.}
\label{PA_lambda}
\end{figure}

\subsection{Parameter Analysis}
\subsubsection{Effect of weight factor $\lambda$}
We now investigate the effect of the weight factor $\lambda$ by plotting the cross-modal retrieval performance versus $\lambda$ on the testing set of XMediaNet in Fig. \ref{PA_lambda}. From the Figure, one could see that both the two components, \ie, semantic alignment ($\mathcal{L}_{PLC}$) and relation alignment ($\mathcal{L}_{BHG}$) are important to improve the robustness against noisy labels, which is consisted with the observation in ablation study. In addition, we can also see that UOT-RCL is not sensitive to the selection of $\lambda$. Specifically, our method can obtain decent performance in a relatively larger range of $\lambda$, \ie, 0.1 $\sim$ 0.7.

\subsubsection{Effect of moving-average factor $\gamma$}
We then explore the effect of training target updating factor $\gamma$ on UOT-RCL performance.
We set $\gamma$ to 0, 0.1, 0.95, 0.99, and 1 to study its influence on retrieval results, where $\gamma = 0$ denotes the target is determined entirely by the prototype embeddings, and $\gamma = 1$ denotes the target will not be updated and thus the cost matrix $\bm{C}_s$ is fixed. Fig. \ref{momentum} shows the learning curves of UOT-RCL on the validation set of Wikipedia.  From the figure, we can see that the sensitivity of $\gamma$ is different in distinct noisy settings. When the noise ratio is 20\%, UOT-RCL can obtain stable performance in different values of $\gamma$, in which a relatively higher $\gamma$ (\ie, 0.95, 0.99, and 1) can bring better results. However, when the noise ratio is 80\%, $\gamma = 1$ can result in overfitting and reduce the model performance. This observation indicates that the cost matrix $\bm{C}_s$ based only on feature-space neighbors is sub-optimal, and it is effective to make $\bm{C}_s$ dynamically updated with a moving-average strategy. Furthermore, the over-reliance on prototype embeddings in updating the target can also lead to sub-optimal results (\ie, $\gamma = 0$ and $\gamma = 0.1$). Overall, setting $\gamma$ as 0.99 can achieve superior performance under different noise ratios.

\begin{figure*}[htbp]
\centering
\subfloat[Original image features]{\includegraphics[width=0.329\textwidth]{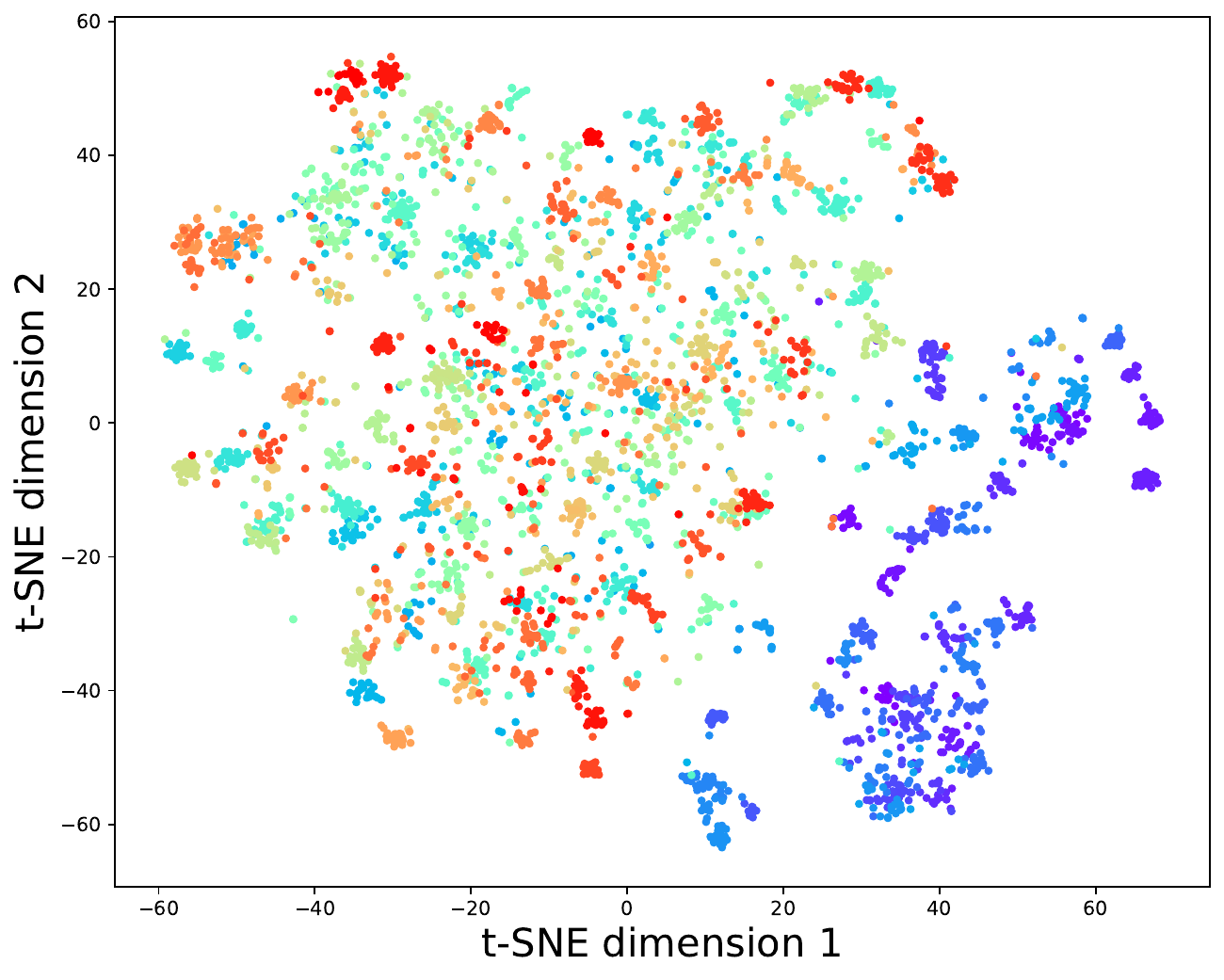}%
\label{tsne_img_ori}}
\subfloat[Original text features]{\includegraphics[width=0.329\textwidth]{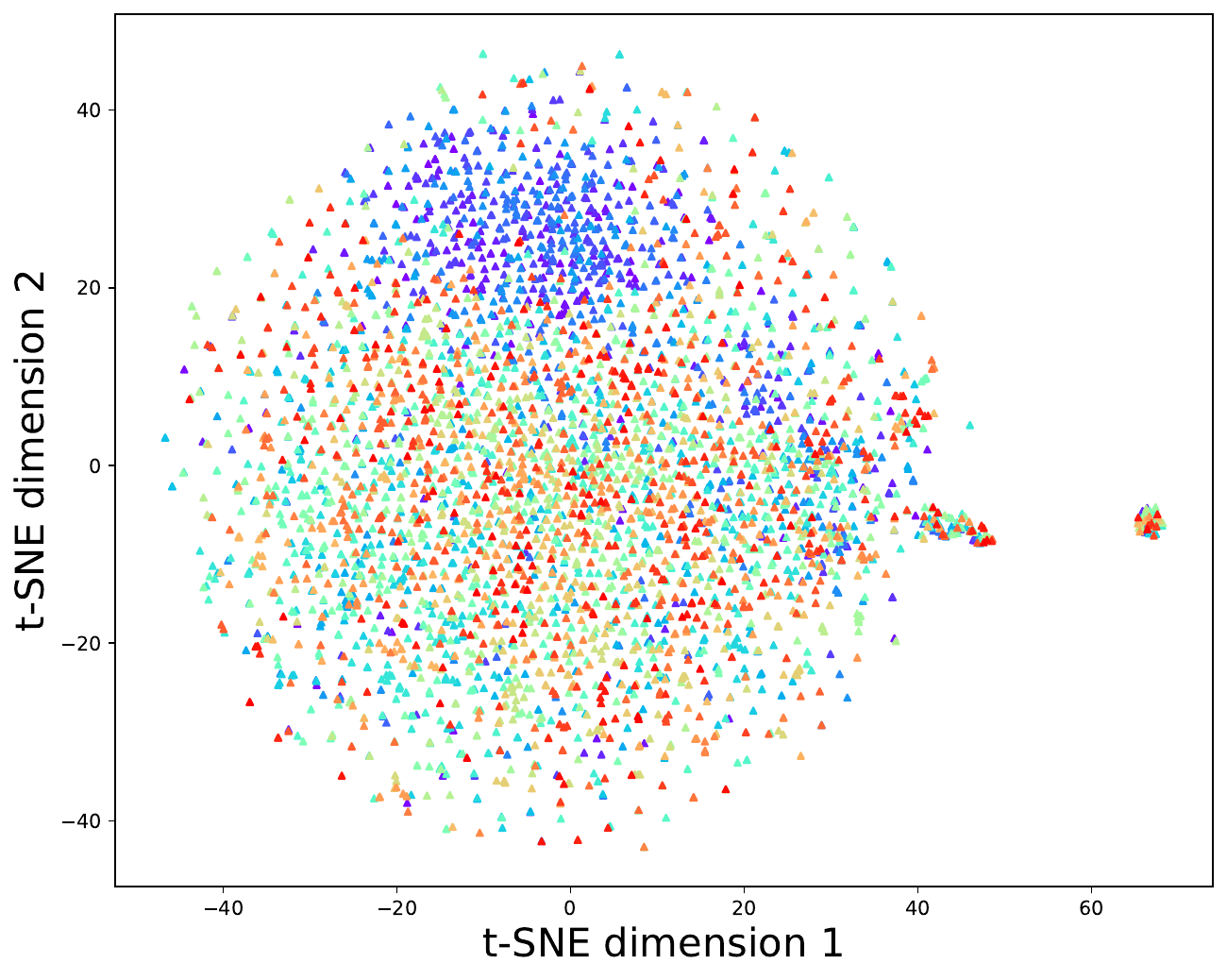}%
\label{tsne_txt_ori}}\hspace{1mm}
\subfloat[Original image-text features]{\includegraphics[width=0.329\textwidth]{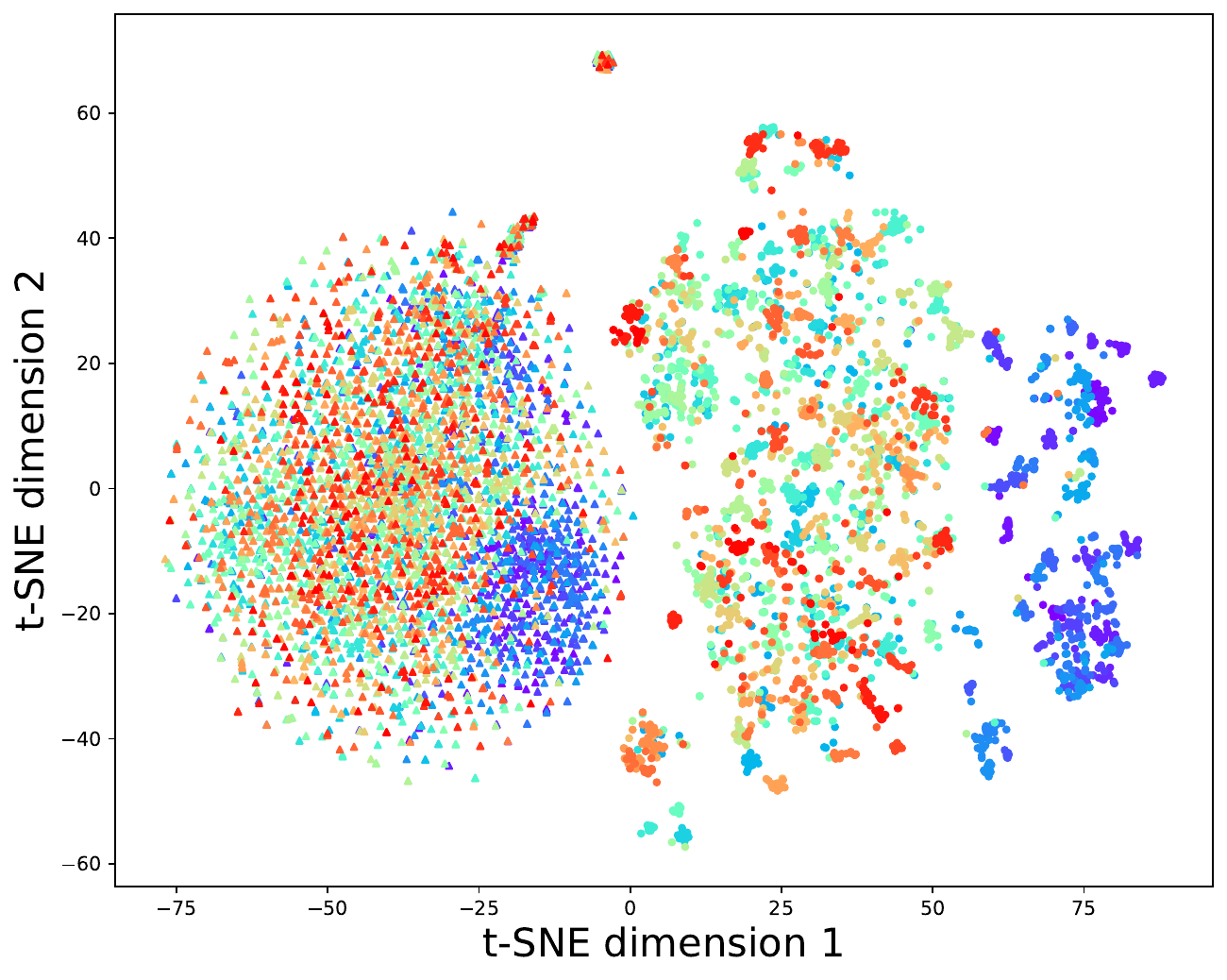}%
\label{tsne_all_ori}}

\vspace{-.35cm}
\subfloat[UOT-RCL image features]{\includegraphics[width=0.329\textwidth]{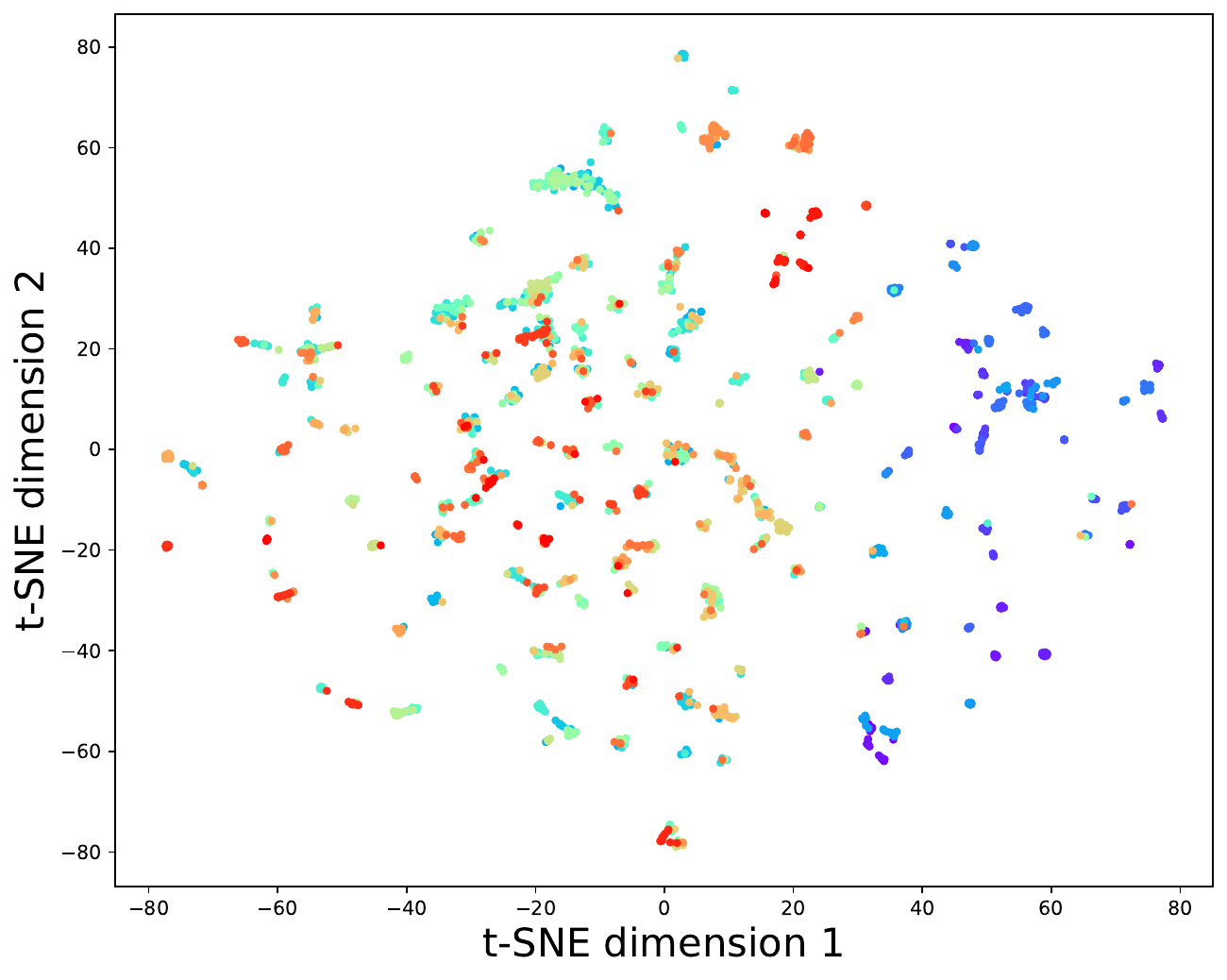}%
\label{tsne_img}}
\subfloat[UOT-RCL text features]{\includegraphics[width=0.329\textwidth]{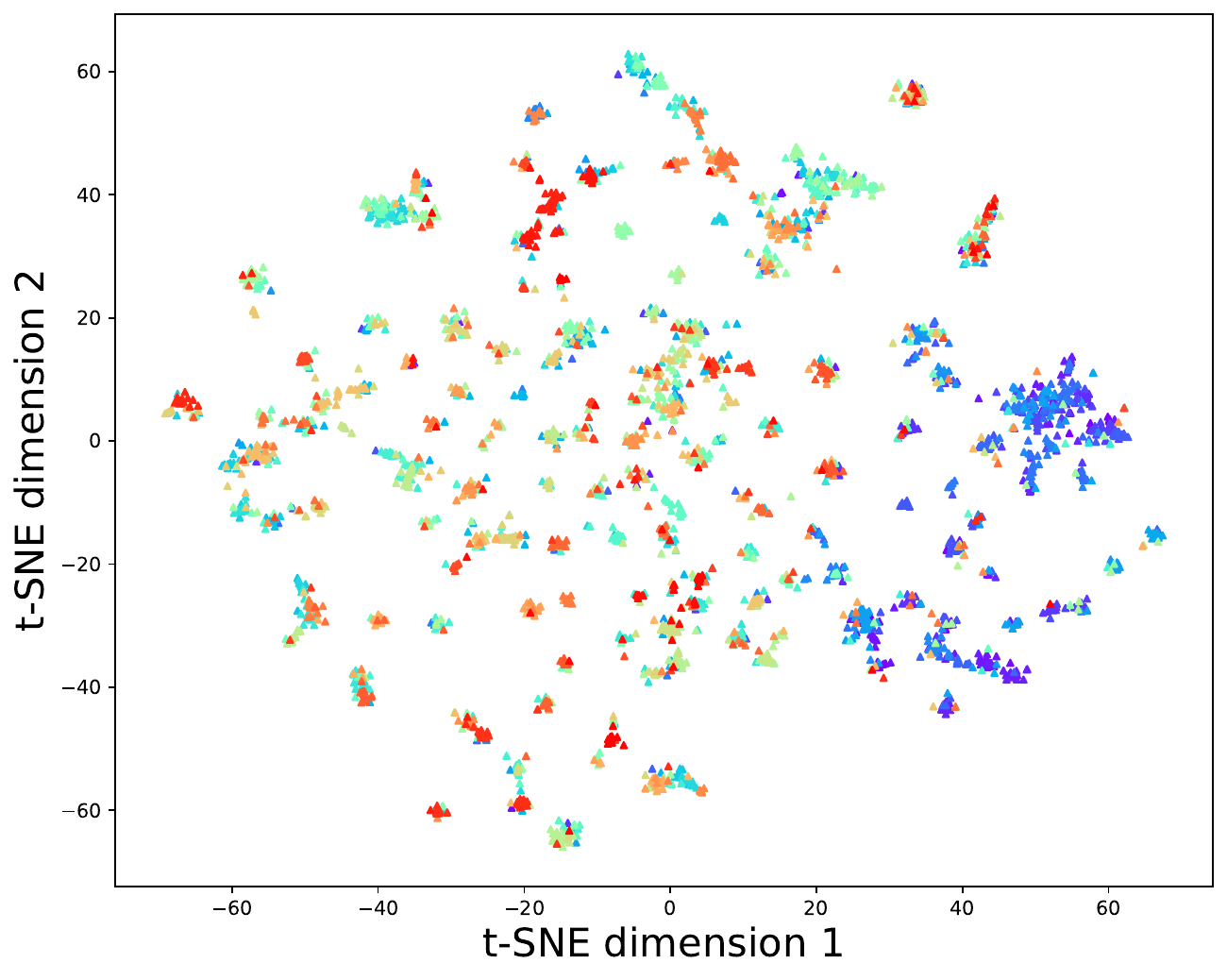}%
\label{tsne_txt}}\hspace{1mm}
\subfloat[UOT-RCL image-text features]{\includegraphics[width=0.329\textwidth]{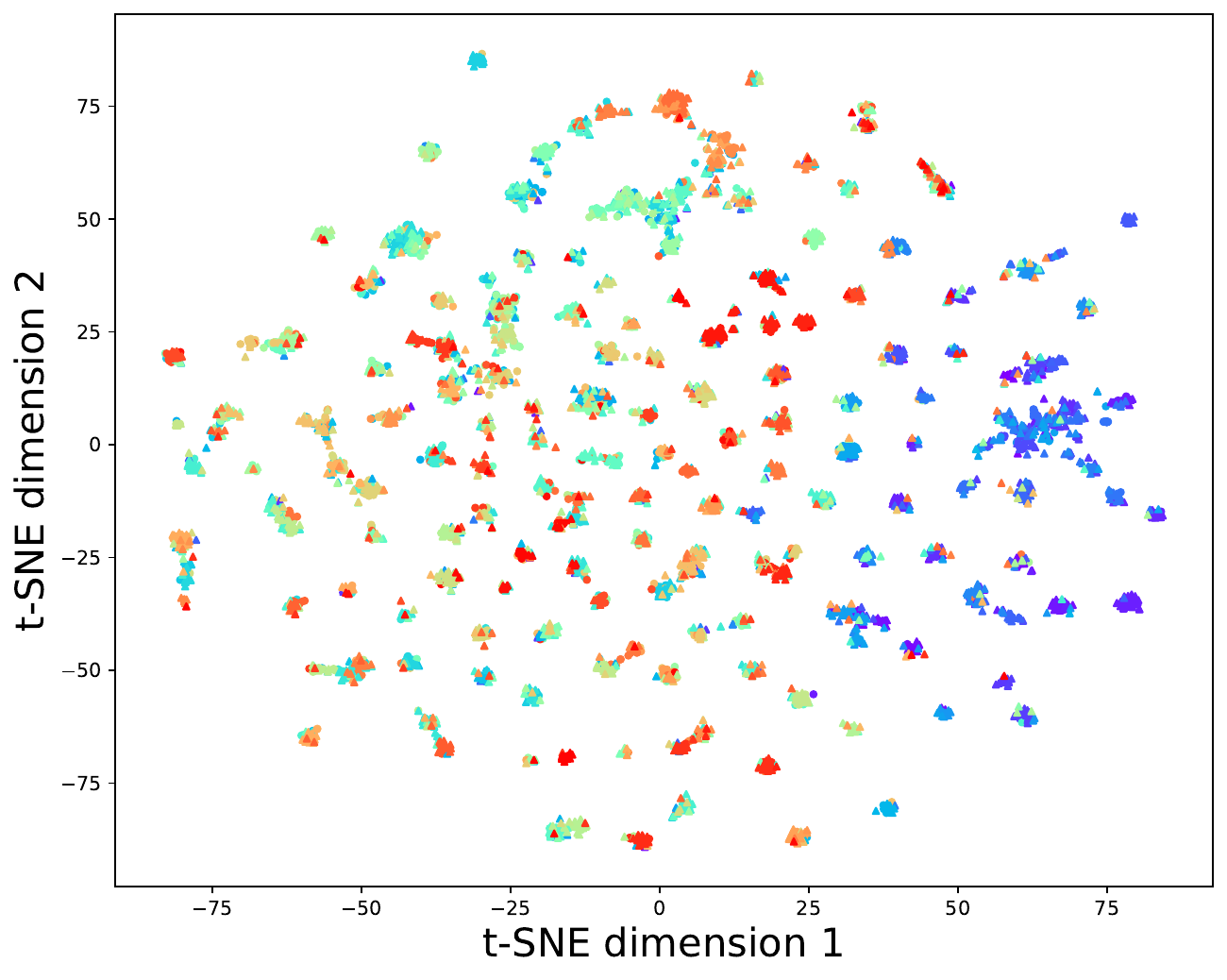}%
\label{tsne_all}}
\caption{The t-SNE visualization on XMediaNet dataset under 20\% noise ratio. Different colors
represent the corresponding classes.}
\label{tSNE}
\end{figure*}

\subsection{Visualization and Analysis}
\subsubsection{Retrieved Examples}
To visually illustrate the retrieval performance of our UOT-RCL, we conduct the case study on the XMediaNet validation set to show some retrieved text and image samples using image queries and text queries, respectively. Specifically, each figure of Figs.\ref{i2t_1}$\sim$\ref{i2t_3} shows a given image query (left) and its top-5 ranked captions (right). Likewise, each figure of Figs.\ref{t2i_1}$\sim$\ref{t2i_2} shows a given caption query (top) and the corresponding top-3 ranked images (bottom). We also highlight the labels of correct retrieval results in green and incorrect ones in red. From these retrieved examples, one could see that our UOT-RCL successfully retrieves most of the relevant items across different modalities. Although some retrieved items do not belong to the same labels as the query data, their semantic content is also correlated with the given queries. For example, the retrieved images in Fig. \ref{t2i_2} are all related to weapons, \ie, "missile", "submarine", and "rocket", which are close to the given query. In summary, these retrieval examples show the ability of our method to capture latent semantics for robust CMR.

\subsubsection{Visualization on the Representations}
We visualize the visual and textual representation produced by the modal-specific networks using t-SNE in Fig .\ref{tSNE}. We contrast the t-SNE embeddings of the original features (Figs .\ref{tsne_img_ori} $\sim$ \ref{tsne_all_ori}) and the learned features (Figs .\ref{tsne_img} $\sim$ \ref{tsne_all}). The visualizations of original features show the heterogeneous gap among modalities that the spatial distribution of visual and textual features are significantly different. Besides, we can observe that both the image and text features of UOT-RCL are distributed regularly in semantics and aligned closely across different modalities, which validates the effectiveness of our method in learning semantic-level discriminative representations and bridging the heterogeneous gap. The visualization also shows some learned features with class overlapping due to the presence of noisy labels, which leads to sub-optimal retrieval performance.

\begin{figure*}[t]
\setlength {\belowcaptionskip} {0cm}
\centering
\subfloat[20\% Noise Ratio]{\includegraphics[width=0.25\textwidth]{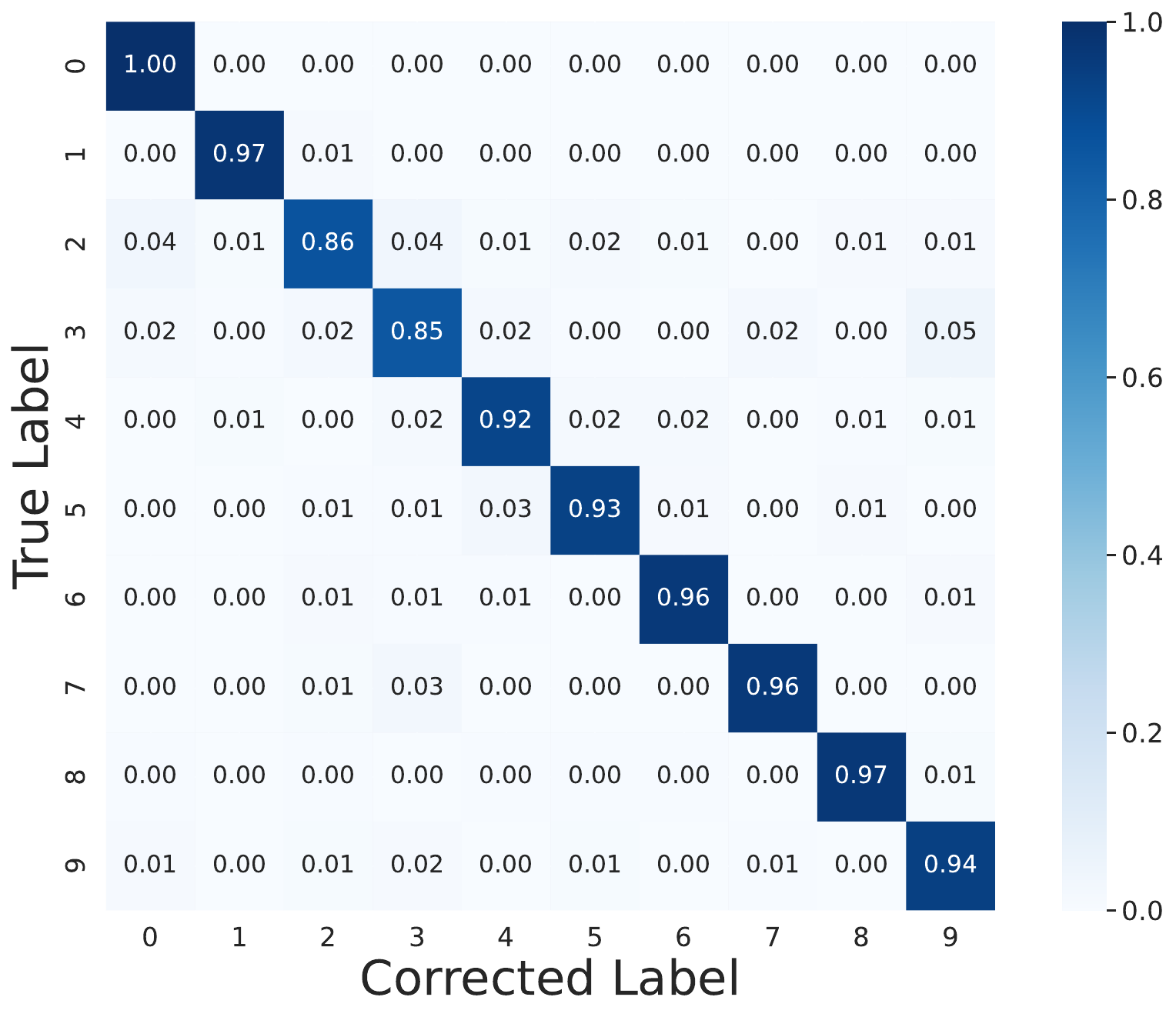}%
\label{vis_2}}
\subfloat[40\% Noise Ratio]{\includegraphics[width=0.25\textwidth]{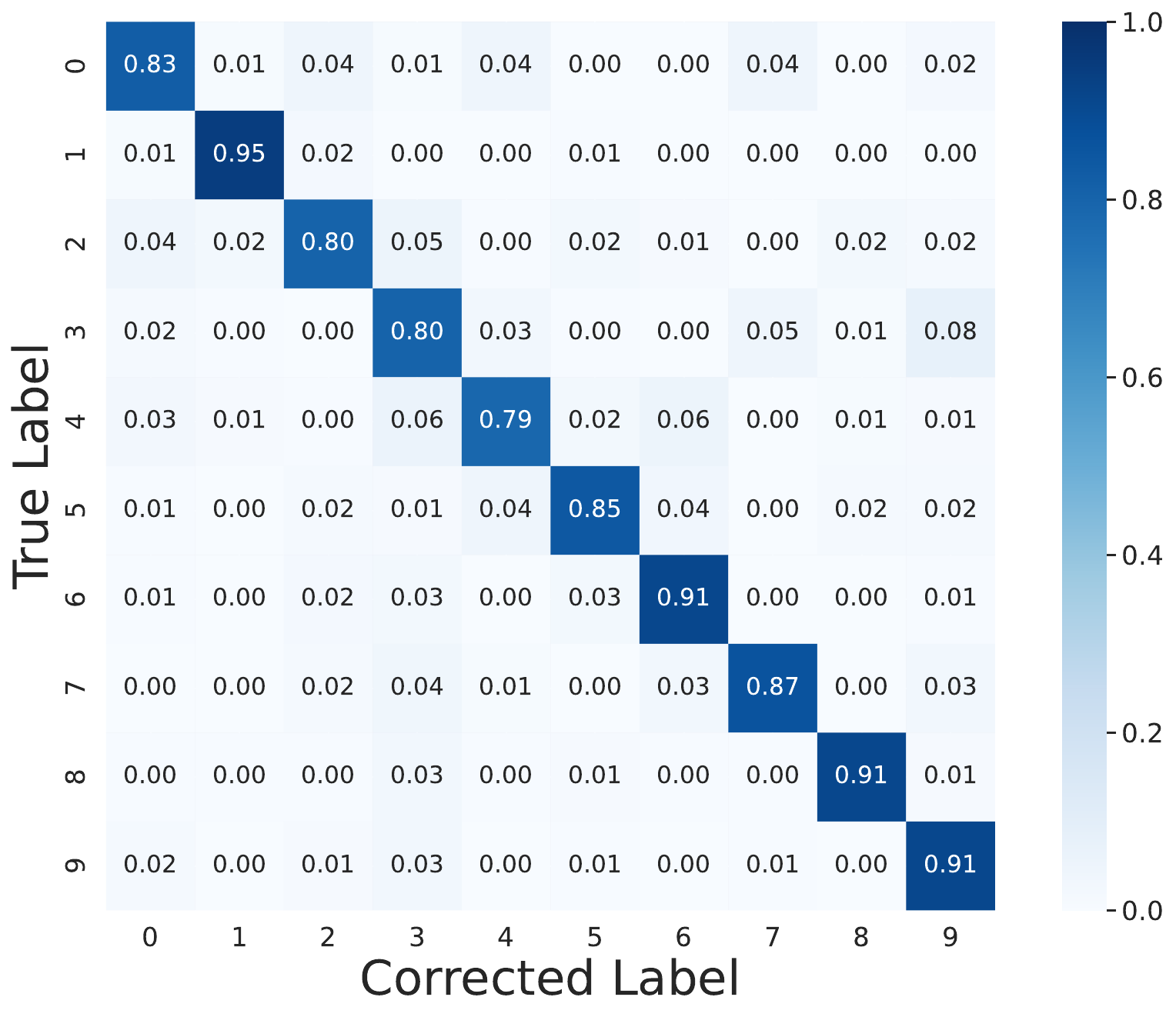}%
\label{vis_4}}
\subfloat[60\% Noise Ratio]{\includegraphics[width=0.25\textwidth]{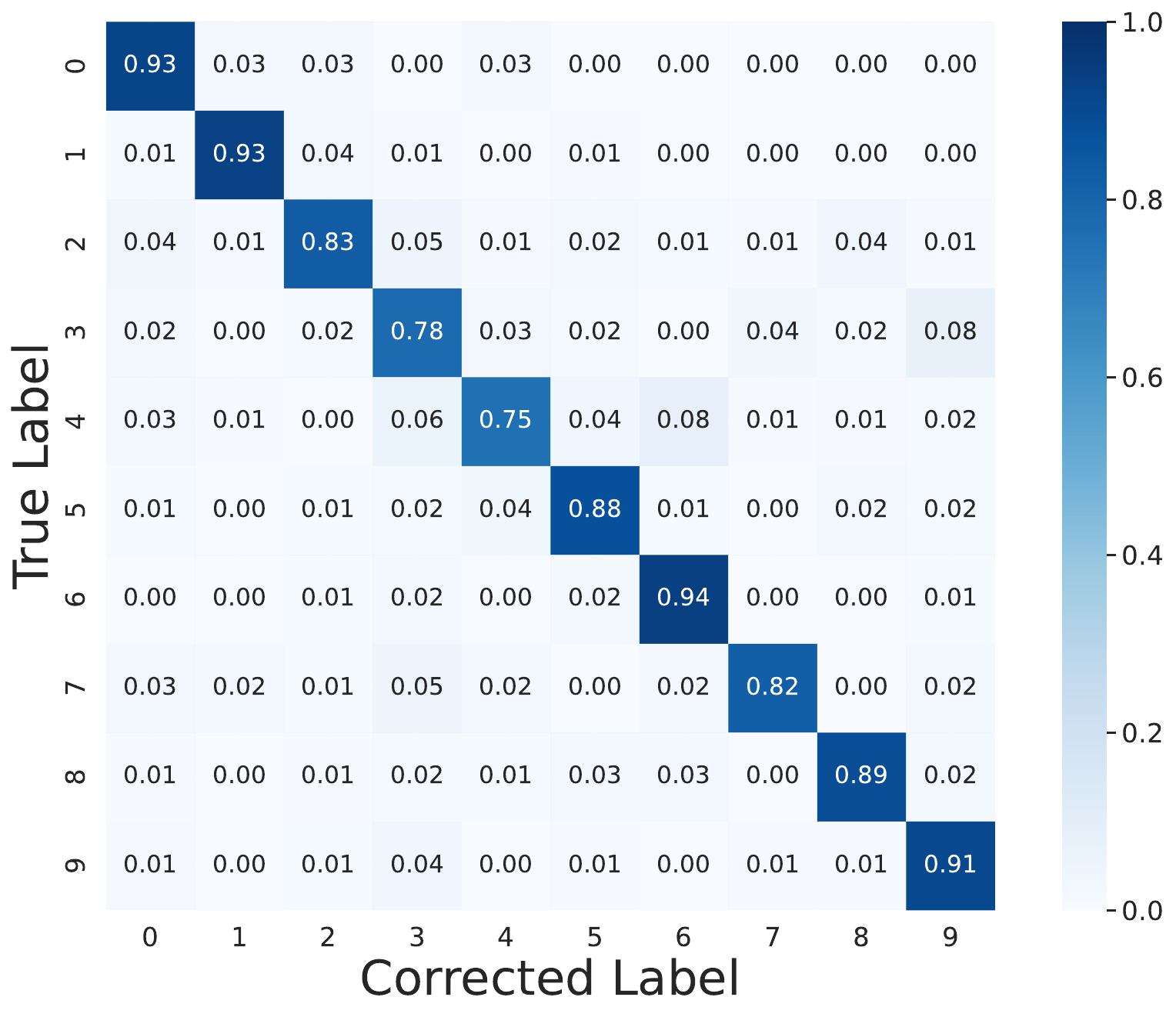}%
\label{vis_6}}
\subfloat[80\% Noise Ratio]{\includegraphics[width=0.25\textwidth]{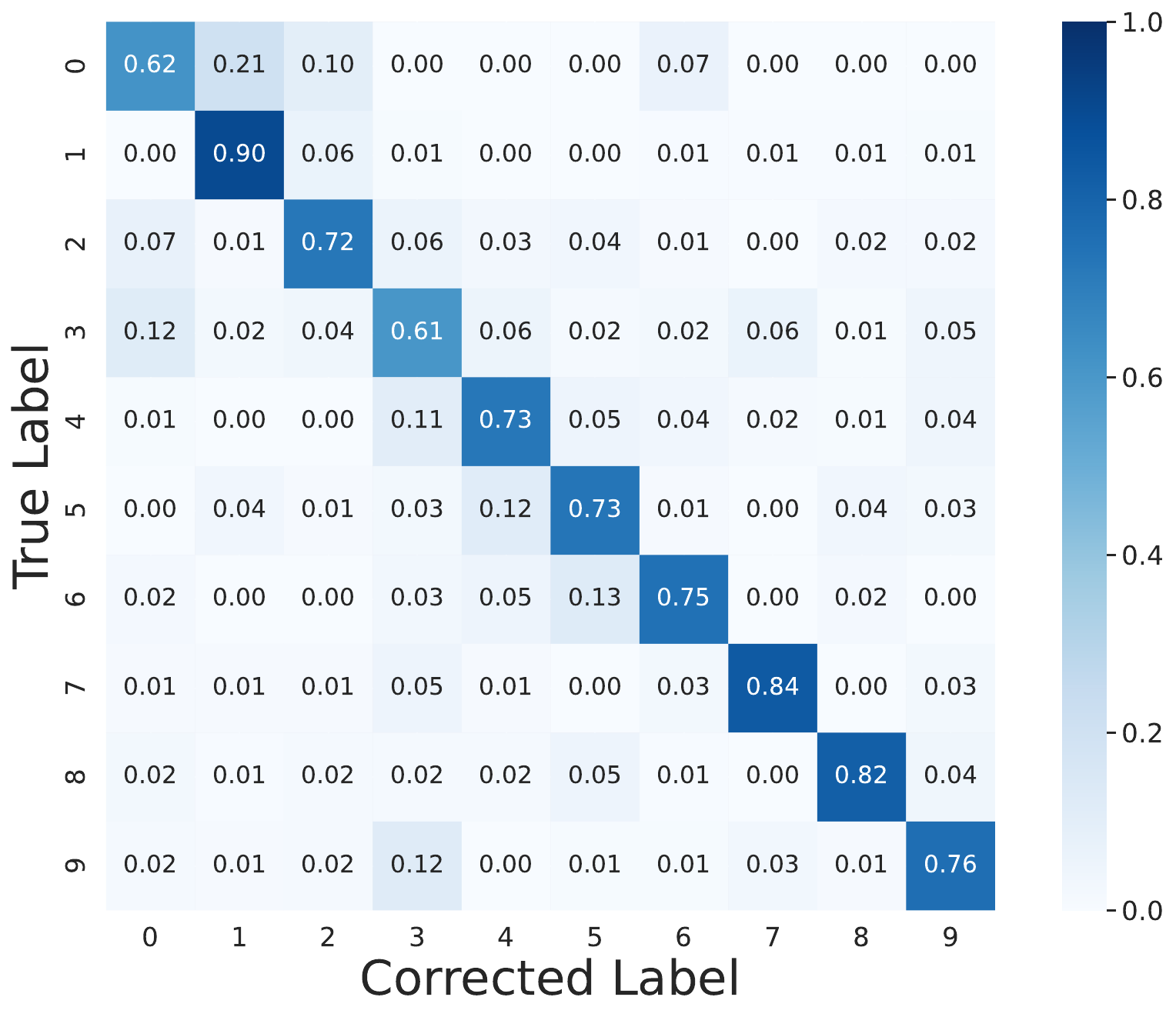}%
\label{vis_8}}
\caption{The ability of our UOT-RCL to correct the noisy labels. The figures show the heatmAP visualization of the probability distribution between our corrected labels and the ground-truth labels under different noise ratios.}
\label{vis}
\end{figure*}

\subsubsection{Visualization on the Corrected Labels}
In this section, we carry out experiments on Wikipedia under different noise ratios to visually analyze the accuracy of the corrected labels generated by our UOT-RCL. Fig.\ref{vis} plots the heatmAP of the probability distribution between the corrected labels and the ground-truth labels on the noisy training set. We can observe that our UOT-RCL successfully generates more reliable corrected labels for all noisy settings. Compared with previous robust methods, \ie, MRL and ELRCMR, trying to down-weight the relative loss for noisy labeled samples, our UOT-RCL goes beyond them by correcting the labels to be reliable, which allows leveraging the training samples more fully.

\section{Conclusion}
In this paper, we propose to use Optimal Transport to handle the noisy labels and heterogeneous gap in cross-modal retrieval under a unified framework. Our UOT-RCL comprises two key components and uses the inherent correlation among modalities to facilitate effective transport cost. First, to combat noisy labels, we view the label correction procedure as a partial OT problem to progressively correct noisy labels, where a novel cost function is designed to provide a cross-modal consistent cost function. Second, to bridge the heterogeneous gap, we formalize a relation-based OT problem to infer the semantic-level cross-modal matching.  Extensive experiments are conducted on several benchmark datasets to verify that our UOT-RCL can endow cross-modal retrieval models with strong robustness against noisy labels.

\small
\bibliographystyle{IEEEtran}
\bibliography{sample-base}

\begin{IEEEbiography}[{\includegraphics[width=1in,height=1.25in,clip,keepaspectratio]{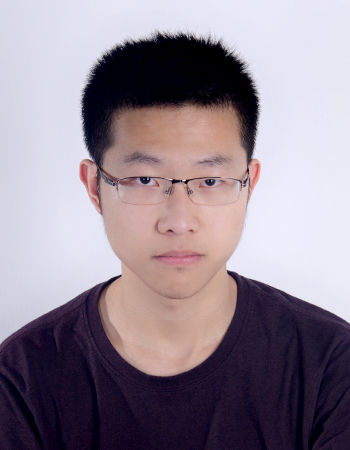}}]{Haochen Han} received the B.S. degree from the School of Energy and Power Engineering, Chongqing University, in 2019. Currently, he is a Ph.D. student in the Department of Computer Science and Technology at Xi'an Jiaotong University, supervised by Professor Qinghua Zheng. His research interests include robust multi-modal learning and its applications, such as action recognition and cross-modal retrieval.
\end{IEEEbiography}
\vskip 0pt plus -1fil


\begin{IEEEbiography}[{\includegraphics[width=1in,height=1.25in,clip]{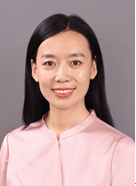}}]{Minnan Luo} received the Ph. D. degree from the Department of Computer Science and Technology, Tsinghua University, China, in 2014. Currently, she is a Professor in the School of Electronic and Information Engineering at Xi’an Jiaotong University. She was a PostDoctoral Research with the School of Computer Science, Carnegie Mellon University, Pittsburgh, PA, USA. Her research focus in this period was mainly on developing machine learning algorithms and applying them to computer vision and social network analysis.
\end{IEEEbiography}
\vskip 0pt plus -1fil

\begin{IEEEbiography}[{\includegraphics[width=1in,height=1.25in,clip]{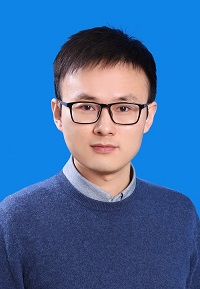}}]
{Huan Liu} received the B.S. and Ph.D. degrees from Xi’an Jiaotong University, Xi’an, China, in 2013 and 2020, respectively. He is currently an Associate Professor with the Department of Computer Science and Technology, Xi’an Jiaotong University. His research interests include affective computing, machine learning, and deep learning.
\end{IEEEbiography}
\vskip 0pt plus -1fil

\begin{IEEEbiography}[{\includegraphics[width=1in,height=1.25in,clip,keepaspectratio]{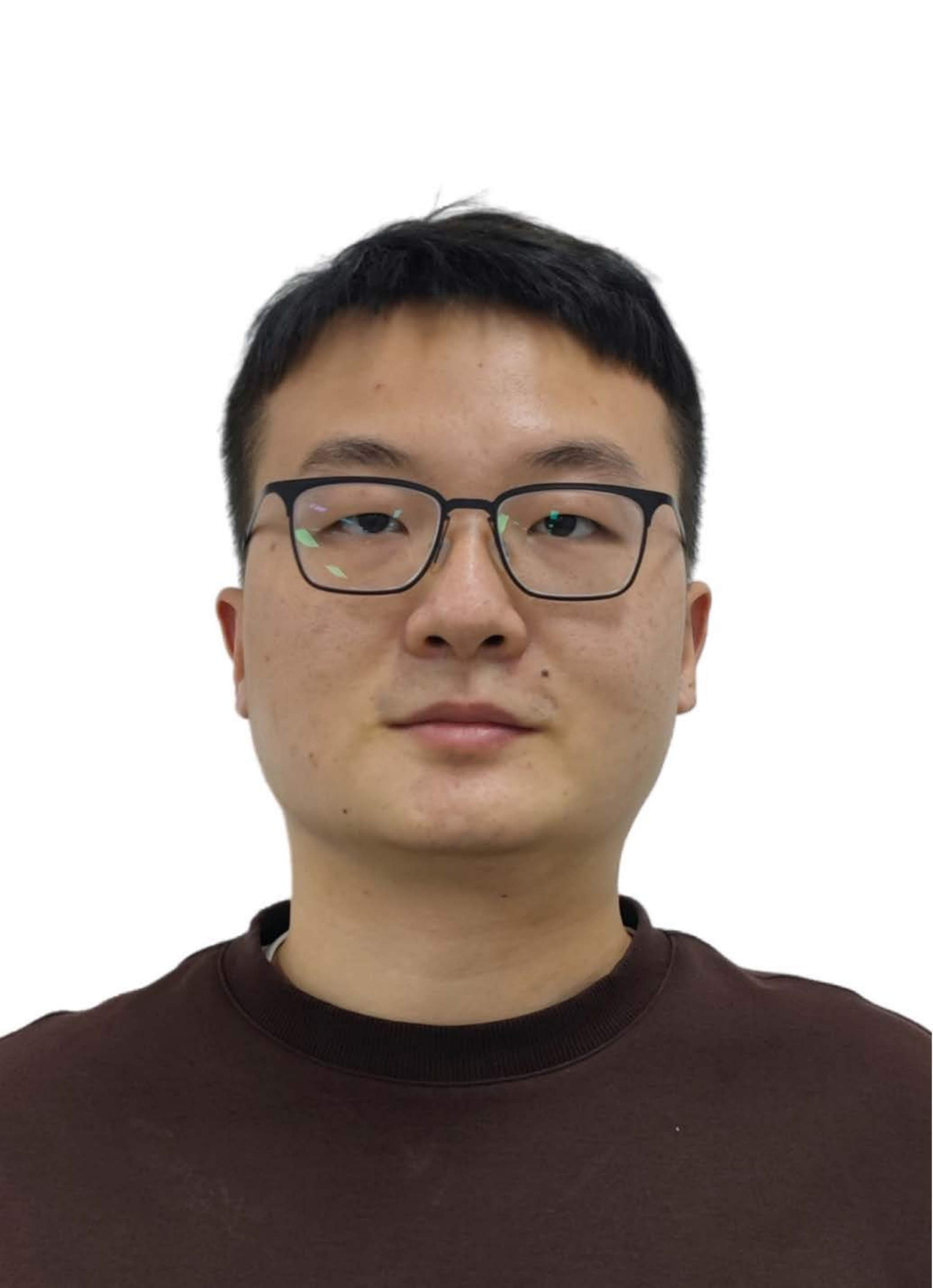}}]{Fang Nan} is a Ph.D. student with Faculty of Electronic and Information Engineering, Xi'an Jiaotong University, Xi'an, China. He received the B.E. degree from Xi'an Jiaotong University in 2017.
His research interests include computer vision and deep learning.
\end{IEEEbiography}
\vspace{-1cm}

\end{document}